\documentclass{article} % For LaTeX2e
\usepackage{floatrow}
\newfloatcommand{capbtabbox}{table}[][\FBwidth]
\newfloatcommand{capbtafig}{figure}[][\FBwidth]

\usepackage{iclr2023_conference,times}
\usepackage{hyperref}

\usepackage{amsmath,amsfonts,bm}

\def\eqref#1{equation~\ref{#1}}
\def\1{\bm{1}}

\DeclareMathAlphabet{\mathsfit}{\encodingdefault}{\sfdefault}{m}{sl}
\SetMathAlphabet{\mathsfit}{bold}{\encodingdefault}{\sfdefault}{bx}{n}

\newcommand{\R}{\mathbb{R}}

\DeclareMathOperator*{\argmax}{arg\,max}

\usepackage{hyperref}
\usepackage{url}
\usepackage{dsfont}
\usepackage{wrapfig}

\def\ie{\textit{i}.\textit{e}., }

\def\eg{\textit{e}.\textit{g}., }
\def\wrt{\textit{w}.\textit{r}.\textit{t}. }

\title{Uncertainty-driven active vision for implicit\\ scene reconstruction}

\author{ 
\And \And \And \And 
  Edward J. Smith$^{1,2}$\thanks{Correspondence to: edward.smith@mail.mcgill.ca} \And
   Michal Drozdzal$^{1}$ \And  \And \And \And \And 
  Derek Nowrouzezahrai$^{1, 2}$\And
   David Meger$^{2}$\And
  Adriana Romero-Soriano$^{1,2}$ \And \\
  \:\:\: \:\:\: \:\:\: \:\:\: \:\:\:\:\:\: \:\:\: \:\:\: \:\:\:  \:\:\:$^1$ Facebook AI Research \:\:\:
$^2$ McGill University \:\:\:
}

\begin{document}

\maketitle
\begin{abstract}
Multi-view implicit scene reconstruction methods have become increasingly popular due to their ability to represent complex scene details. Recent efforts have been devoted to improving the representation of input information and to reducing the number of views required to obtain high quality reconstructions. Yet, perhaps surprisingly, the study of which views to select to maximally improve scene understanding remains largely unexplored. We propose an uncertainty-driven active vision approach for implicit scene reconstruction, which leverages occupancy uncertainty accumulated across the scene using volume rendering to select the next view to acquire. To this end, we develop an occupancy-based reconstruction method which accurately represents scenes using either 2D or 3D supervision. We evaluate our proposed approach on the ABC dataset and the in the wild CO3D dataset, and show that: (1) we are able to obtain high quality state-of-the-art occupancy reconstructions; (2) our perspective conditioned uncertainty definition is effective to drive improvements in next best view selection and outperforms strong baseline approaches; and (3) we can further improve shape understanding by performing a gradient-based search on the view selection candidates. Overall, our results highlight the importance of view selection for implicit scene reconstruction, making it a promising avenue to explore further.\looseness-1
\end{abstract}

\section{Introduction}

Recent advances leveraging implicit neural representations have dramatically increased the capacity for scene understanding~\citep{mescheder2019occupancy, park2019deepsdf}. 
For example, in the space of neural rendering, a large number of works have focused on devising new methods to better understand scenes with only 2D supervision~\citep{yu2021pixelnerf, yariv2020multiview}, with their 
 widespread adoption in part due to their ability to express far more complex scenes and details than explicit counterparts such as meshes or voxels ~\citep{wang2018pixel2mesh, xie2019pix2vox}. Moreover, given the potential of these functions to recover scene properties such as geometry, lighting and object semantics, they hold the promise of revolutionizing applications in augmented reality, autonomous driving and robotics. Current state-of-the-art models, however, may require up to hundred views to achieve high quality scene reconstructions. Some efforts have been devoted to drastically reduce this number by leveraging dataset amortization 
 ~\citep{yu2021pixelnerf, schwarz2020graf, chen2021mvsnerf}. Surprisingly though, no work has focused on studying the effect of active view selection on the scene reconstruction quality under a small budget constraint -- \eg using up to five views.

Active view selection methods aim to manipulate the viewpoint of a camera to choose views that best improve 3D scene understanding ~\citep{connolly1985determination}. These methods have traditionally leveraged heuristics such as maximal coverage~\citep{pito1999solution} or information gain~\citep{krainin2011autonomous} without data priors to iteratively select the next best view to acquire over depth images and the shape they directly provide. In the context of contemporary reconstruction methods, without levering data, heuristics like coverage or information gain cannot distinguish between unobserved regions of the scene and regions for which a reconstruction model is \emph{uncertain}, and as a result acquire views which may provide no additional scene understanding. Recently we have witnessed the introduction of data-driven approaches to learn next best view policy models~\citep{vasquez2021next} as well as 3D shape reconstruction models to drive these policies \citep{yang2018active}. These data-driven models often require additional information such as 3D shape supervision~\citep{yang2018active} or depth to train~\citep{peralta2020next}, which is rarely available, difficult to obtain and no longer a requirement for any aspect of scene reconstruction when leveraging neural implicit functions~\citep{yu2021pixelnerf, chen2021mvsnerf}.

\begin{wrapfigure}{r}{7.2cm}
\capbtafig{%
\scalebox{.9}{
  \vspace{0.2cm}
  \scalebox{1}{
        \centering
        \includegraphics[trim={0cm 0 0cm 0},clip, width=1\linewidth]{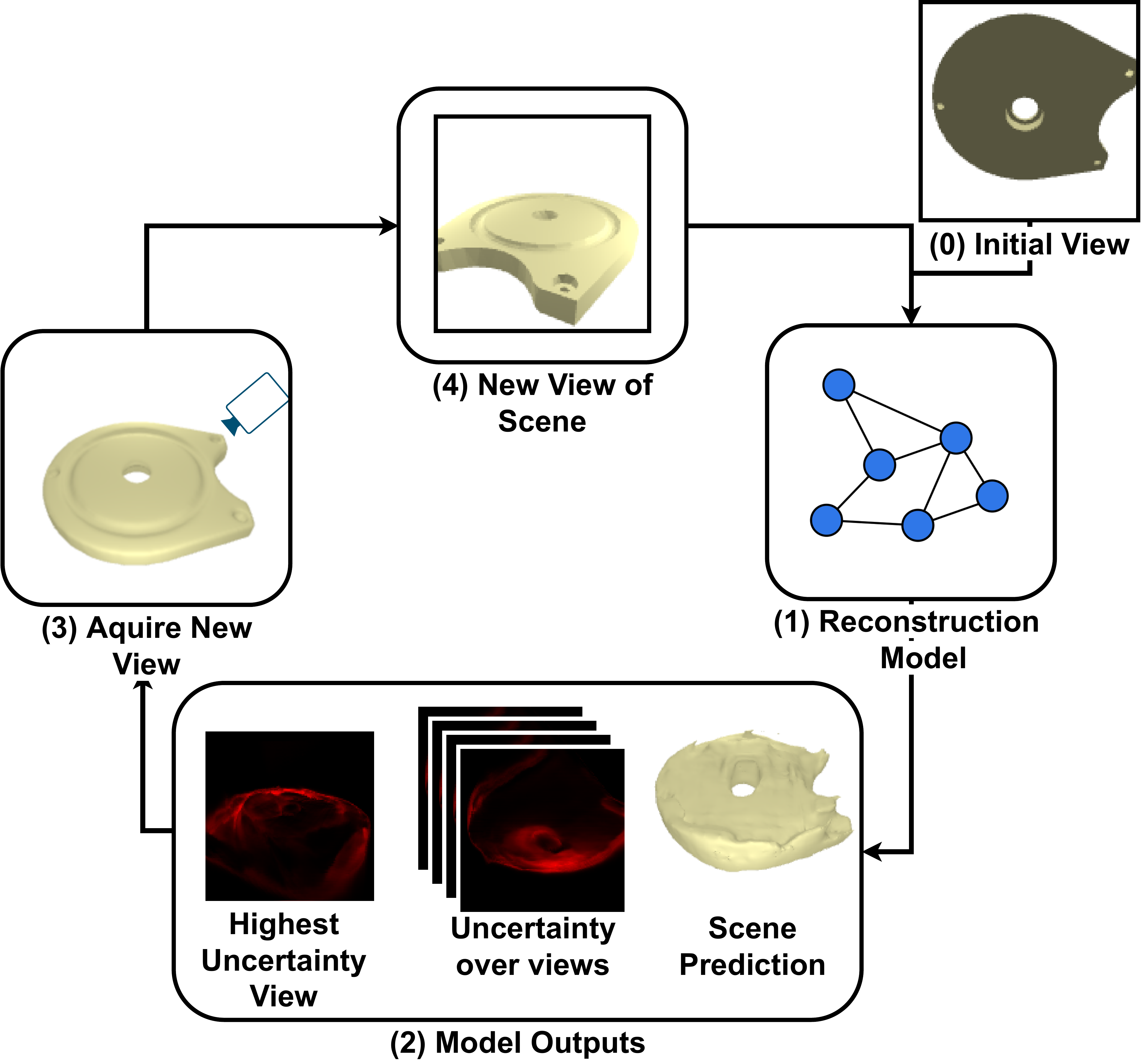}
  }
  }
  }
  {
\caption{Uncertainty-driven active vision pipeline.\vspace{-0.3cm}}

    \label{fig:method_overview}
}
\end{wrapfigure}
We propose an \emph{uncertainty-driven} active vision approach whose goal is to choose the sequence of views which lead to the highest reduction in reconstruction uncertainty (see Figure~\ref{fig:method_overview}). The proposed approach, %called \emph{XXXX}, 
introduces an implicit multi-view reconstruction model to predict occupancy, leverages the occupancy predictions to estimate uncertainty over unseen views, and defines view selection policies which seek to maximize the observable model uncertainty. Notably, the contributed reconstruction model is robust to arbitrary numbers of input views, and can be trained by leveraging either full 3D supervision from occupancy values or 2D supervision from renderings. Moreover, the observable model uncertainty is estimated by extending the volume rendering formulation to accumulate predicted occupancy probabilities along rays cast into the scene, enabling the search space of possible views to be efficiently explored. 
We evaluate our proposed active vision approach on the simulated ABC~\citep{Koch_2019_CVPR} dataset as well as the challenging, in the wild CO3D~\citep{reizenstein2021common} scene dataset, by leveraging up to $5$ image perspectives. Our results demonstrate that: (1) our reconstruction model obtains impressive reconstructions which lead to visible improvement over the previous state-of-the-art multi-view occupancy method on the ABC dataset, and perhaps surprisingly, this improvement persists even when training with only 2D supervision for larger numbers of input views; (2) our uncertainty-driven active vision approach achieves notable improvements in shape understanding under volumetric and projection-based metrics relative to strong baseline selection policies on both ABC and CO3D datasets; and (3) by performing a gradient-based search on the view selection candidates, we can further improve shape understanding. The code to reproduce the experiments is provided:  \href{https://github.com/facebookresearch/Uncertainty-Driven-Active-Vision}{https://github.com/facebookresearch/Uncertainty-Driven-Active-Vision}.

\section{Related Works}

Traditional active vision methods for 3D reconstruction, limited by lack of access to contemporary learning methods and large scale data, generally focus on identifying views to maximize visibility of unobserved areas of the scene using a range camera~\citep{pito1999solution, connolly1985determination, banta2000next}. \cite{connolly1985determination} first proposed to determine views in the scene which would maximize the visibility of unobserved voxels. Many works then focused on reducing the cost of computing coverage metrics and increasing the number of candidate views considered~\citep{pito1999solution,  blaer2007data, low2006adaptive, vasquez2013hierarchical}. Conversely, other methods computed utility scores over additional factors such as view overlap, scan quality and navigation distance which can be optimized over to select views~\citep{massios1998best, fisher1999next, foissotte2008next, vasquez2014volumetric}. More contemporary next best view methods, especially in the context of robotics, focused on maximizing information gain as opposed to direct view coverage~\citep{sebastian2005probabilistic, le2008active, huber2012bayesian, krainin2011autonomous, peng2020viewpoints}, though this optimization was over models without strong data priors. The absence of data-driven reconstruction models in these methods results in depth information being necessary for both reconstruction and view selection, coarse shape predictions relative to learning based approaches, and view selections which cannot reason over learned shape priors.  

A small number of active vision methods have also been proposed which make use of deep learning. \cite{mendoza2020supervised} trained a deep learning classifier to predict which pose out of a set of discreet options will best improve a generated point cloud, and~\cite{vasquez2021next} regressed the pose for a camera which would maximize coverage, though both operate over a range camera with no learned reconstruction model. Most similar to our setting, \cite{peralta2020next} used reinforcement learning to select optimal paths for an RGB camera, over a pre-trained reconstruction algorithm. This involved the training of a reinforcement learning policy on top of the reconstruction algorithm, both of which required ground truth shape depth information for training, whereas our method requires no additional learning and can be applied with only 2D supervision. \cite{yang2018active} proposed a data-driven recurrent reconstruction method with a unified view planner, though the voxel predictions here are coarse and the learning is performed on-policy, with their reconstruction model biased towards views selected by their policy, and so not directly comparable. Finally, in the similar setting of active haptic perception, \cite{smith2021active} learn where to touch an object next to best understand shape over a learned data-driven reconstruction model, also using reinforcement learning approaches.

\section{Method}
In our active vision approach, the goal is to choose a sequence of views which lead to the highest reduction in reconstruction uncertainty, and as result improve the 3D shape reconstruction accuracy. An overview of the proposed pipeline is depicted in Figure~\ref{fig:method_overview}: (1) a pre-trained shape reconstruction model is fed with an object image; (2) the predicted reconstruction is used to estimate the uncertainty over the unseen object views; (3) the view with the highest uncertainty is acquired and subsequently fed to the reconstruction model, which is designed to process an arbitrary number of views. We begin by describing our proposed reconstruction model, which can be trained by leveraging either full 3D supervision from occupancy values or 2D supervision from renderings. Afterwards, we will present our proposed uncertainty-driven next best view selection approach.

\begin{figure}
    
 \centering
 \includegraphics[width=.8\textwidth]{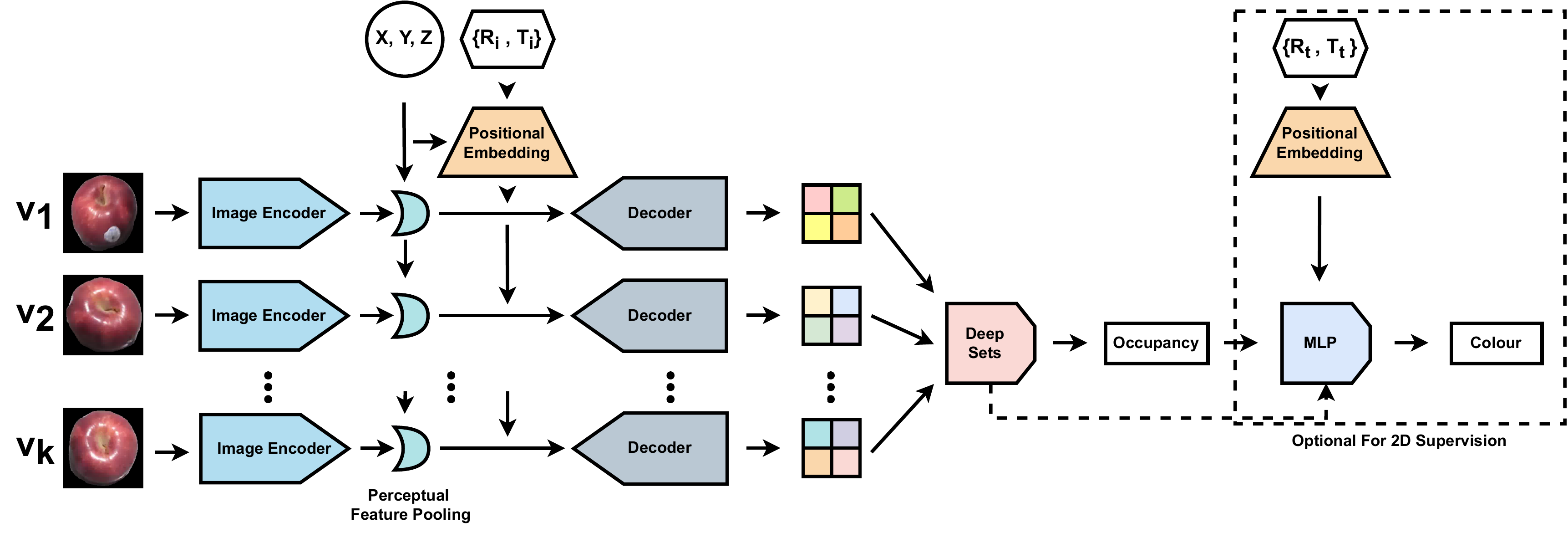}
 \caption{Our reconstruction method. $(X, Y, Z)$ is the input 3D position in space, $\{R_i, T_i\}$ is the set of input image camera parameters, and  $\{R_t, T_t\}$ is the target camera parameters. }
 \label{fig:network_diagram}
\end{figure}

\subsection{Reconstruction Model}
Our proposed reconstruction model, depicted in Figure~\ref{fig:network_diagram}, is robust to arbitrary numbers of input views and produces occupancy predictions. The model takes as input a position in space, $(X, Y, Z)$, a set of $K$ of input images $v_i$ and their corresponding camera parameters $(R_i, T_i)$, and produces an occupancy prediction. In particular, an image encoder extracts features from each input image through a large VGG-like CNN \citep{simonyan2014very} followed by a perceptual feature pooling operation~\citep{wang2018pixel2mesh}. Then, the features of each image are concatenated with a positional embedding~\citep{mildenhall2020nerf} of their corresponding camera parameters and the input position, and are passed through a series of ResNet Blocks~\citep{he2016deep}. The resulting camera-position-aware features of each image are aggregated using deep set pooling layers~\citep{zaheer2017deep}, allowing for permutation invariant aggregation of features from arbitrary numbers of views. Finally, a sigmoid activation is applied to produce an occupancy prediction. The reconstruction model is trained using full 3D supervision from ground truth occupancy values through a combination of intersection over union (IoU) and binary cross entropy (BCE) losses.

We extend the model to operate without full 3D supervision and to leverage 2D supervision from renderings. In this case, the model takes the camera parameters of a target view $(R_t, T_t)$ as additional input. These parameters are embedded \citep{mildenhall2020nerf} and concatenated with the predicted occupancy and intermediate features from the deep set camera-position-aware feature aggregation. The result of the concatenation is passed through a series of fully connected layers to predict colour for the target position. The reconstruction model is trained by leveraging a rendering loss. Along a ray $r= r_o + td$, where $r_o$ is the ray origin and $d$ is the ray viewing direction, we compute a colour value along a ray $\hat{C}(r)$ by integrating occupancy $o(t)$ and colour $c(t)$ predictions: 
\begin{equation}
     \hat{C}(r) = \int^{t_f}_{t_n} T(t)o(t)c(t) dt,
\label{eq:VR}
\end{equation}
where $t_n$ and $t_f$ define the range of integration, and $T(t) = exp(-\int^t_{t_n} o(s)ds)$ allows accumulation of colour up to occlusions \citep{oechsle2021unisurf}. The models are trained by randomly selecting between 1 and 5 input images and minimizing the mean squared error (MSE) between predicted pixel values along rays and ground truth pixel values in a target image. Further training details are provided in the Appendix and an architecture diagram for the model is provided in Figure \ref{fig:network_diagram}.
 
\begin{figure}[t]
    \centering
    \includegraphics[width=0.8\linewidth]{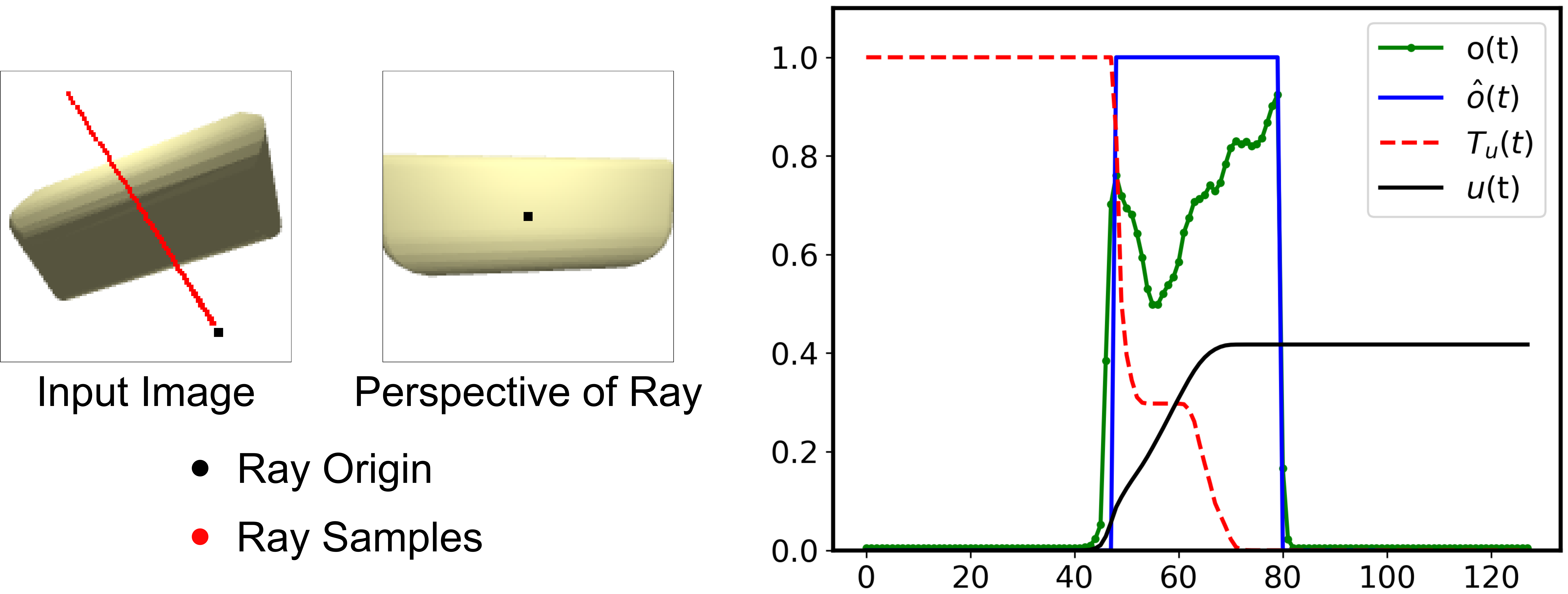}
    \caption{Demonstration of the accumulation of uncertainty along a ray. On the left we display the initial input image for to the model and in the middle a new perspective into the scene for a ray, with the ray origin labeled in black, and samples along the ray labeled in red. In the graph on the left we highlight the per-sample ground truth occupancy values $\hat{o}(t)$, predicted occupancy values $o(t)$, and resulting uncertainty accumulation function values $T_u(t)$ and accumulated uncertainty $u(t)$. }
    \label{fig:main_ray}
\end{figure}

\subsection{Uncertainty-driven next best view selection}
% \subsection{Binary Classification Uncertainty}
We start by defining occupancy uncertainty from occupancy predictions. Then, we introduce our proposed view uncertainty computation, and finally we present the uncertainty-driven policies considered for the task of next best view selection.

\subsubsection{Occupancy uncertainty}
Occupancy prediction, as a binary classification task, is determined by applying a threshold to predicted network probabilities. For our tasks, we set the threshold for predictions at 0.5. The distance of predicted probabilities from this decision boundary provides implicit model confidence \citep{mlbook, wu2018sharing, zhou2012self}. Scaling this value by two provides a normalized confidence score for model predictions: $2|0.5 -o(t)|$; however, it is well known that binary classification probabilities are poorly calibrated with accuracy \citep{guo2017calibration}, and so we calibrate this confidence score using an exponential : $(2|0.5 -o(t)|)^\beta$, where $\beta \in \R^{+}$ is a hyper-parameter which either smooths or exaggerates the distances from the decision boundary (see Figure \ref{fig:calibration}). We then define the uncertainty of occupancy prediction in as follows: 
\begin{equation}
    u(o(t)) = 1 - (2 | o(t) - .5 | ) ^{\beta}.
    \label{eq:occ_uncert}
\end{equation}
In the Appendix we demonstrate our occupancy predictions are initially quite poorly calibrated, and that by identifying the correct value of $\beta$ we drastically improve calibration error. 

\iffalse
\begin{figure}
\begin{floatrow}
\capbtafig{%
\scalebox{.8}{
  \vspace{0.2cm}
  \scalebox{1}{
        \centering
        \includegraphics[width=1\linewidth]{Figures/Method/Calibration.png}
    }
    }
    }
    {
    \centering
     \caption{ Confidence scores for occupancy predictions under different calibration hyperparameter settings.}
 \label{fig:calibration}
    }

\end{floatrow}
\end{figure}
\fi
\begin{minipage}{.33\textwidth}
  \centering
 \includegraphics[width=1.\linewidth]{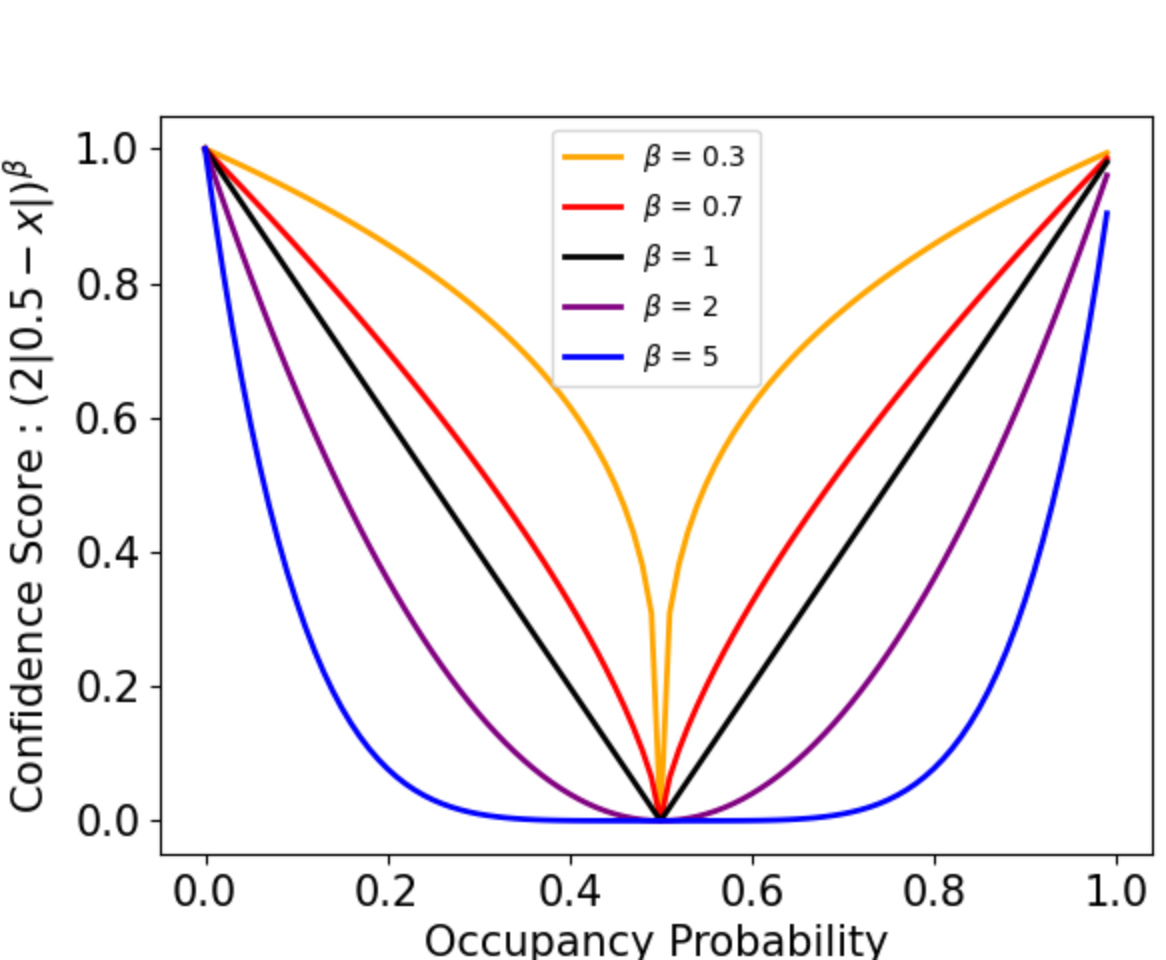}
 \vspace{-0.3cm}
  \captionof{figure}{Occupancy confidence for different calibration hyper-parameter values.}
  \label{fig:calibration}
\end{minipage}%
\hspace{0.2cm}
\begin{minipage}{.63\textwidth}
  \centering
 \includegraphics[width=1.\textwidth]{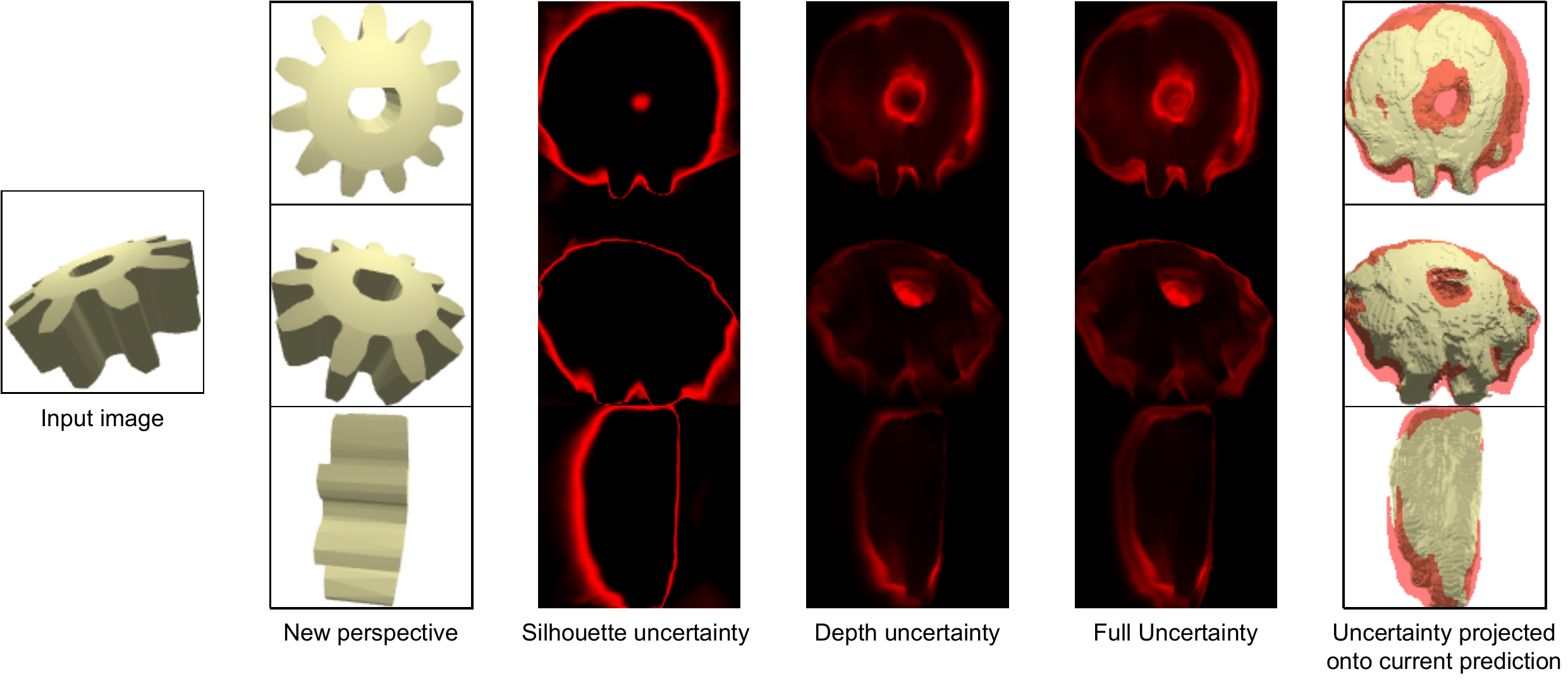}
  \captionof{figure}{ Visualization of uncertainty sources (cols. 3-5) for 3 new views (col. 2) after an initial view (col. 1), with high view uncertainty projected onto the current prediction (col. 6).}
  \label{fig:uncert_view}
\end{minipage}

\subsubsection{View uncertainty}
We decompose the observable uncertainty from a given perspective (view) into two sources: \emph{silhouette} uncertainty and \emph{depth} uncertainty. Intuitively, for a given ray through a scene, the uncertainty associated with silhouette prediction addresses the question ``does this ray hit an object?'', and if the ray has been established to hit an object, the uncertainty associated with the depth of occlusion addresses the question ``where does this ray first hit an object?''. 

\textbf{Silhouette Uncertainty.} A ray's silhouette prediction, $s(r) \in [0, 1] $, can be resolved using the final accumulation value $T(t_n)$ as follows: $s(r) = 1 - T(t_n)$. 
As this is an occupancy prediction, we leverage Equation~\ref{eq:occ_uncert} to define silhouette uncertainty, $u_{sil}(r)$, as follows: 
$
    u_{sil}(r) = 1 - (2 | s(r) - .5 | ) ^{\lambda_s}.
$\looseness-1

\textbf{Depth Uncertainty.} For the occupancy of a 3D point in space, we again leverage Equation ~\ref{eq:occ_uncert}: 
$
    u_{p}(t) = 1 - (2 | o(t) - .5 | ) ^{\lambda_u}.
$
Then, we seek to accumulate point uncertainty along a ray. In particular, we aim to accumulate uncertainty indiscriminately up until the model is highly confident that a surface has been observed, as this point represents where depth uncertainty has been fully resolved. We therefore update the accumulation function $T(t)$ in Equation~\ref{eq:VR} to account for uncertainty as follows: $T_u(t) = \mathrm{exp}(-\int^t_{t_n} \mathds{1}_{s} |o(s)-.5|^{\lambda_t} ds)$, where $\mathds{1}_{s}$ is the indicator function for $o(s) >0.5$, and $\lambda_t \in \R^{+}$ is a smoothing hyper-parameter. This ensures the accumulation of uncertainty along the ray reduces with the degree to which the model is positively confident an occlusion has been observed  -- \ie $o(t) > .5$.

With an uncertainty definition over the scene and an accumulation function to integrate them up to occlusions, we possess the minimum tools to apply volume rendering. However, we also consider that our model predictions are limited in resolution which may lead to high uncertainty regions at decision boundaries regardless of shape understanding. 

To mitigate this issue, we introduce a rate of change correction, $d(t) = 1 - (\nabla_r o(t))^{\lambda_d}$, where $\nabla_r o(t) \in [0,1]$ is the directional derivative of the occupancy prediction along the ray $r$, and $\lambda_d \in \R^{+}$ is a smoothing hyper-parameter. Multiplying $u_{p}(t)$ by $d(t)$ reduces the defined uncertainty at a point when passing through a tight surface decision boundary, as defined by the rate of change of the occupancy. 

We rewrite the volume rendering definition highlighted in Equation \ref{eq:VR} to account for depth uncertainty along a ray as:  
\begin{equation}
    u_{depth}(r) = \int^{t_f}_{t_n} T_u(t)  d(t) u_{p}(t) dt.
\end{equation}
In Figure \ref{fig:main_ray}, we highlight how predicted occupancy values along a ray through a scene results in the accumulation of uncertainty. In this example uncertainty is present due to the model's lack of confidence in the exact location of the object's outer shell.  

\textbf{View Uncertainty.} We define the uncertainty of a perspective as the average accumulated silhouette and depth uncertainty from rays cast from it: 
\begin{equation}
    u(v) = \frac{1}{|v|}\sum_{r\in v} (u_{sil}(r) + \lambda) * u_{depth}(r),
\end{equation}
where $\lambda$ is a hyper-parameter which ensures that depth uncertainty is present even if silhouette uncertainty is zero. The uncertainty under this combination of sources is highlighted in Figure \ref{fig:uncert_view}, which demonstrates high uncertainty both in areas where the model is unsure of the existence of the shape along a ray (silhouette uncertainty) and location of the object along a ray (depth uncertainty). We provide an ablation over our uncertainty definition in the Appendix. 

\subsubsection{Next best view selection policies}

We consider two uncertainty-driven policies for next best view selection, with the first inspired by traditional discrete next best view selection policies \citep{pito1999solution, connolly1985determination}, and the second allowing for more efficient large scale scene exploration. 

\textbf{Candidate Policy.} This policy considers a set of $N$ random views, $V = \{v_i\}_{i=1}^{N}$, chosen such that no views are closer than $\delta$ to each other. The policy selects the view which has the largest computed view uncertainty: $\argmax_{v_i \in V} u(v_i)$ . 

\textbf{Gradient Policy.} This policy selects a random view from the search space of views such that the selected view is no closer than $\delta$ to any views already selected. Then, the position of this view is updated using $m$ steps of gradient ascent to maximize the uncertainty it contains. This optimization is performed with a distance penalty to push the updated position away from the positions of already sampled views, and to encourage it to stay within the search space of views: 
\begin{equation}
    \max_{\theta}  u(v_{\theta}) - \lambda_D D(v_{\theta}, \{v_{<i}\}) - \lambda_S D(v_{\theta}, S), 
\end{equation}
where $D$ is the $l_2$ distance metric, $\lambda_D$ and  $\lambda_S$ are hyper-paramters,  $\{v_{<i}\}$ is the set of previous views, $S$ is the defined search space, and $\theta$ is the viewing parameters of the initial view $v$.

\section{Experiments}
We will now evaluate the performance of our proposed reconstruction method and then compare the proposed uncertainty based next best view policies to a set of strong baselines under two reconstruction settings -- \ie leveraging full 3D supervision and leveraging 2D supervision from renderings. Additional experimental details and hyper-parameter settings can be found in the Appendix.\looseness-1

\textbf{Datasets.} (1) \underline{\textit{ABC dataset}}: We make use of a subset of the ABC dataset \citep{Koch_2019_CVPR} curated by \cite{smith2021active} for haptic active perception. This dataset consists of 25000 object models, split into training, validation and test sets with sizes 23000, 1500, and 500. The perspective search space for this dataset is defined as a sphere around the object where the camera is always pointed towards the center of the object, and the camera resolution is $128\times128$. 
(2) \underline{\textit{CO3D dataset}}: We make use of the CO3D dataset which is made up of $\sim1.5$M images from 19,000 in the wild videos sequences of 50 object classes \citep{reizenstein2021common}. We sub-sample this dataset to only use video instances with more then 10 frames, use the supplied image masks to remove background colours, and scale all images to $256\times256$.  We split the resulting dataset into train, validation and test sets with a ratio of 80/10/10. The perspective search space for each object is the set of images remaining in the video frames after 10 evenly spaced test images, and the set of frames already sampled have been removed. 

\textbf{Next best view selection baselines.} (1) \underline{\textit{Random}}: A random policy which selects views at random but no closer than $\delta$ to existing views. (2) \underline{\textit{Even}}: A uniform coverage policy which selects the view which maximizes the distance from existing views. (3) \underline{\textit{Odd}}: Uniform coverage policy which selects every second view chosen by the \textit{Even} policy, because many objects considered are symmetrical, and as a result the \textit{Even} policy often provides redundant information at every second selection. For all baselines and uncertainty-based policies, the first view is selected at random. 

\textbf{Metrics.} (1) \underline{\textit{IoU}}: we consider the 3D intersection over union metric (IoU) to measure the volumetric performance of our reconstruction models and view selection policies \citep{zhou2019iou}. (2) \underline{\textit{PSNR}}: we consider the peak signal to noise ratio metric (PSNR) to measure the rendering performance of our reconstruction models and view selection polices \citep{hore2010image}.

\begin{figure}
\begin{floatrow}
\capbtabbox{%
\scalebox{.9}{
  \scalebox{0.76}{
    \begin{tabular}{cccccc}
    \toprule
    \multicolumn{1}{c}{} & \multicolumn{5}{c}{\textbf{Number of Images}}  \\
     \textbf{Model} & 1 & 2 & 3 & 4 & 5 \\
    \cmidrule[1pt](lr){1-1}\cmidrule[1pt](lr){2-6}
   Ours - 3D  & \textbf{0.6537} &  \textbf{0.7374} &  \textbf{0.7776} &  \textbf{0.8013} &   \textbf{0.8149} \\
    Ours - 2D  & 0.3601 &  0.4728 &  0.5287 &  0.5675 &  0.5928 \\
   
    \cmidrule[1pt](lr){1-1}\cmidrule[1pt](lr){2-6}
    Pix2Vox & 0.4673 & 0.4919 &  0.5025 &  0.5066 & 0.5087 \\
    \bottomrule
 \end{tabular}
 }
}}{%
\centering
  \caption{Reconstruction IoU results on ABC.}
  \label{table:Recon_3D_main}
  %\small
}
\capbtabbox{%
\scalebox{.9}{
  \scalebox{0.76}{
     \begin{tabular}{cccccc}
    \toprule
    \multicolumn{1}{c}{} & \multicolumn{5}{c}{\textbf{Number of Images}}  \\
     \textbf{Policy} & 1 & 2 & 3 & 4 & 5 \\
    \cmidrule[1pt](lr){1-1}\cmidrule[1pt](lr){2-6}
   Candidate Policy & 17.40 & \textbf{18.43} & \textbf{19.03} & \textbf{19.40} & \textbf{19.61}\\
   Random Policy & 17.40 & 18.37 & 18.92 & 19.30 & 19.53\\
   
    Even Policy  & 17.40 & 18.39 & 18.98 & 19.36 & 19.60\\
    Odd Policy & 17.40 & 18.41 & 19.00 & 19.35 & 19.59\\
    \bottomrule
 \end{tabular}
 }
}}{%
    \centering
  \caption{PSNR comparison over selection policies in the 2D supervision setting on CO3D.}
  \label{table:ablation_CO3D}
  \small
}

\end{floatrow}
\end{figure}

\subsection{Scene Reconstruction Results}
We first evaluate our reconstruction model in the full 3D supervision setting. To do so, we train our model on the ABC dataset and demonstrate its performance by comparing it in terms of IoU to Pix2Vox \citep{xie2019pix2vox}, a state of the art multi-view occupancy reconstruction model. We train Pix2Vox in the same setting as our own model, and show performance from 1 to 5 views in Table~\ref{table:Recon_3D_main}. We highlight object reconstruction examples in the Appendix. We then evaluate our reconstruction model in the 2D supervision setting. We train the model on the ABC dataset and highlight its IoU performance in Table~\ref{table:Recon_3D_main} and provide PSNR results in the Appendix. Relative to Pix2Vox we observe considerable improvement in reconstruction performance in the 3D supervision setting, and perhaps surprisingly, our model in the 2D supervision setting achieves significantly better IoU predictions when more than 3 views are provided. We additionally train our proposed reconstruction model on a subset of the CO3D dataset~\citep{reizenstein2021common} and highlight example reconstruction results in Figure~\ref{fig:3D_comparision_Co3D} and provide PSNR results in the Appendix. From this training we observe strong qualitative results and consistent improvement in PSNR with additional views, though poor silhouette labels and highly variable camera distances in the dataset leads to poor occupancy predictions for certain objects.\looseness-1

\begin{figure}
    \centering
    \vspace{-.4cm}
    \includegraphics[width=\linewidth]{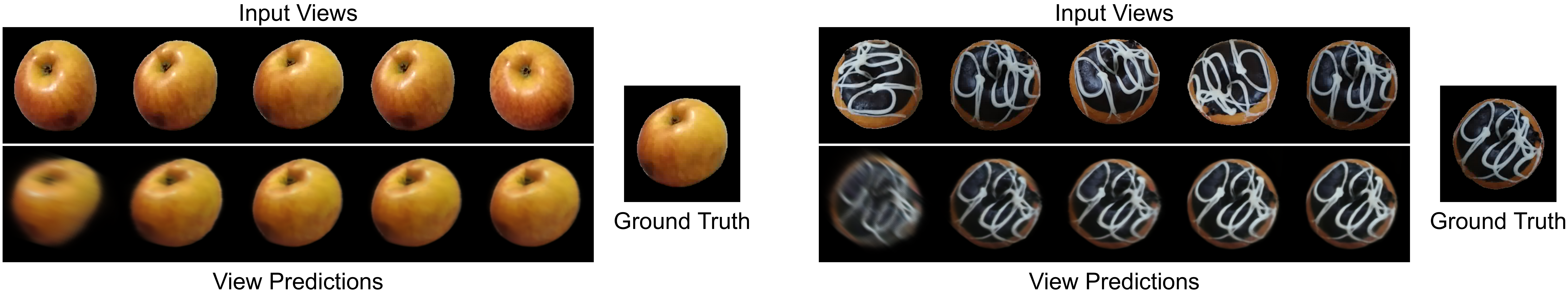}
    \caption{Shape reconstruction examples from the CO3D dataset where $i$-th image in the bottom row represents the reconstruction using input images $1$ to $i$.   }
    \label{fig:3D_comparision_Co3D}
\end{figure}

\vspace{0.1cm}
\begin{minipage}{0.3\textwidth}
    \includegraphics[trim={5mm 3mm 3mm 5mm},clip, width=1\linewidth]{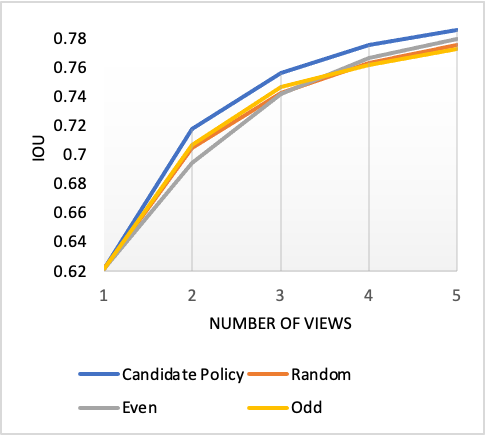}
    \captionof{figure}{IoU comparison of policies -- 3D supervision on ABC.}
    \label{fig:IoU_3D}
\end{minipage}\hfill
\begin{minipage}{0.3\textwidth}
    \includegraphics[trim={5mm 2mm 3mm 5.5mm},clip, width=1\linewidth]{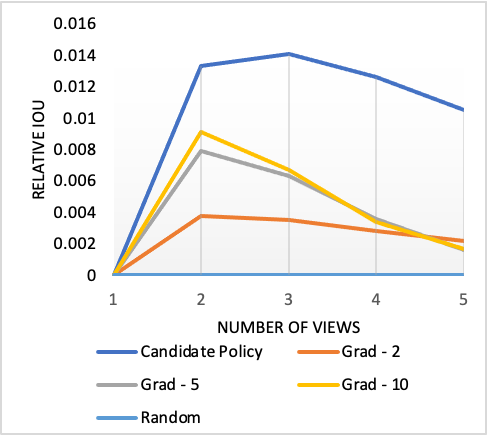}
    \captionof{figure}{IoU improvement over random policy -- 3D supervision on ABC.}
    \label{fig:IoU_3D_Grad}
\end{minipage}\hfill
\begin{minipage}{0.3\textwidth}
    \includegraphics[trim={5mm 2mm 3mm 5mm},clip, width=1\linewidth]{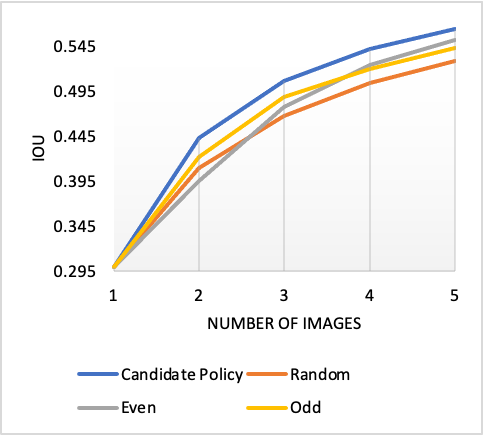}
    \captionof{figure}{IoU comparison of policies -- 2D supervision on ABC.}
    \label{fig:IoU_2D}
\end{minipage}

\subsection{Next Best View Results}
We evaluate the performance of our active vision pipeline by computing uncertainty along a ray as follows: (1) We randomly sample 1024 rays from the set of pixels in each perspective to compute the uncertainties for each view. (2) We estimate the integrals for computing uncertainties along each ray using 128 evenly distributed samples. (3) The directional derivative in the rate of change correction at ray sample $t_i$, $d(t_i)$, is approximated by the absolute change in occupancy in the surrounding samples along the ray: $\nabla_r o(t_i) = | o(t_{i+1}) - o(t_{i-1})|$. All hyper-parameters are chosen using a grid search on a validation set for the downstream next best view task. 
 
\begin{figure}[t]
    \centering
    \includegraphics[width=.9\linewidth]{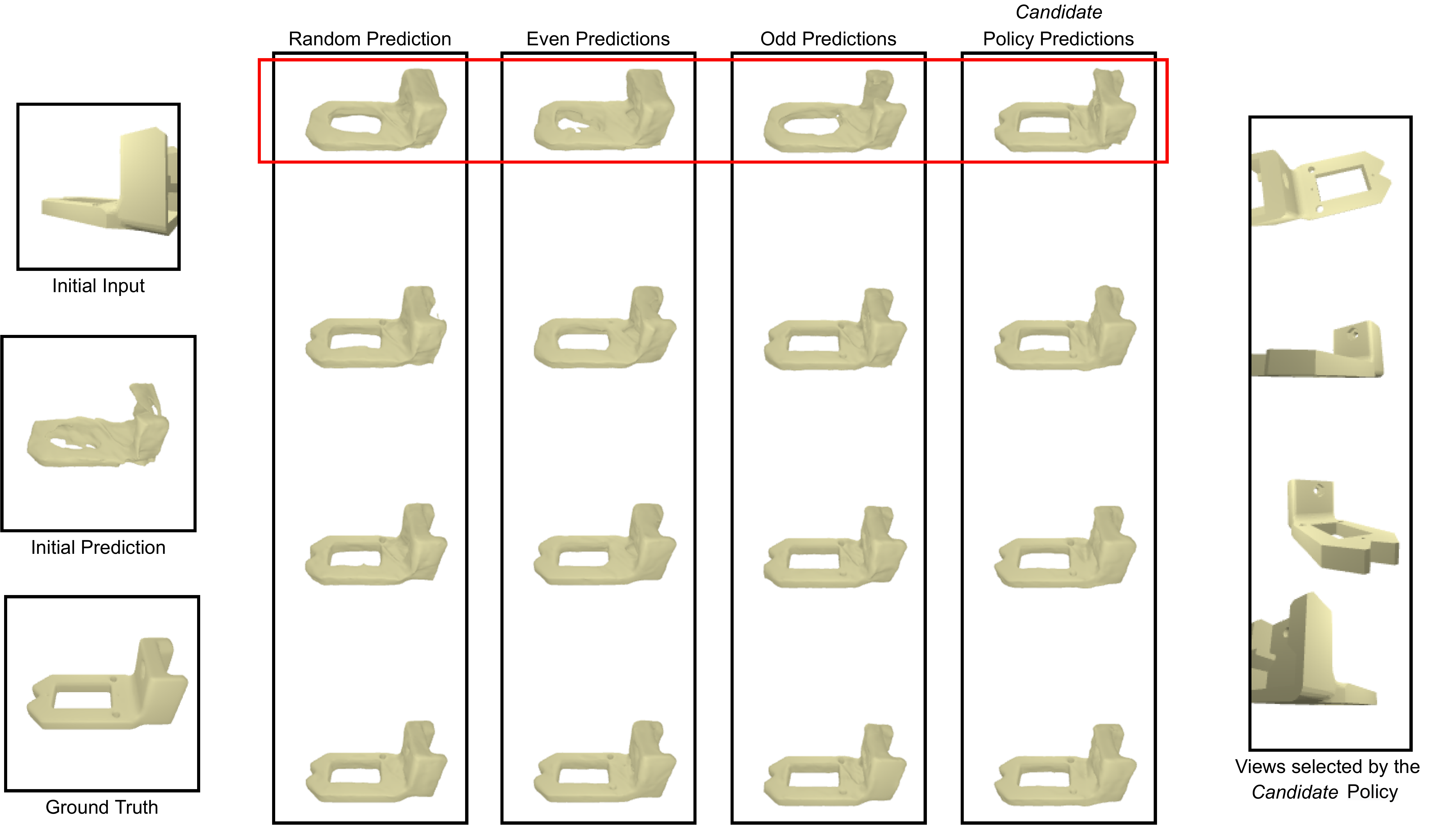}
    \caption{Shape reconstruction comparison, on the ABC dataset in the 3D supervision setting,  over selection policies with the middle columns of images highlighting the reconstructions with 2 to 5 views under each policy and the \textit{candidate} policy view selections highlighted in final column. }
    \label{fig:3D_comparision}
\end{figure}

\textbf{3D Supervision - ABC:} We compare the performance of our \textit{candidate} policy with $20$ candidate views to the baseline policies on ABC. We evaluate using IoU over the test set, with $10$ random view initializations for each object and up to 5 input views. Results are presented in Figure~\ref{fig:IoU_3D}. We observe our policy leads to notably more accurate shape predictions than the baseline policies with slightly diminishing returns as all policies converge to uniform coverage. We highlight example reconstructions in Figure~\ref{fig:3D_comparision}, where our proposed policy recovers both coarse shape and fine grained details with more accuracy and less number of views -- \eg see the reconstruction obtained after the first next best view selection (top row). In the Appendix we show the \textit{candidate} policy selection also results in lower variance and minimum IoU over the $10$ random view initialization. We further evaluate the performance of our \textit{gradient} policy with $2$, $5$ and $10$ gradient steps under the same conditions and compare to the \textit{candidate} policy and the \textit{random} policy with which they are initialized. Figure~\ref{fig:IoU_3D_Grad} shows the relative improvement in performance of these policies \wrt the random policy. The gradient policy significantly outperforms the random policy with only 2 steps, and increasing the number of steps improves this difference but with diminishing returns. However, the gradient policy does not improve over the performance of the candidate policy, as this policy approximately fully covers the perspective search space, while the gradient policy tends to get stuck in local uncertainty maxima. Further candidate and gradient policy results are highlighted in the Appendix.\looseness-1

\begin{figure}[t]
    \centering
    \includegraphics[width=.9\linewidth]{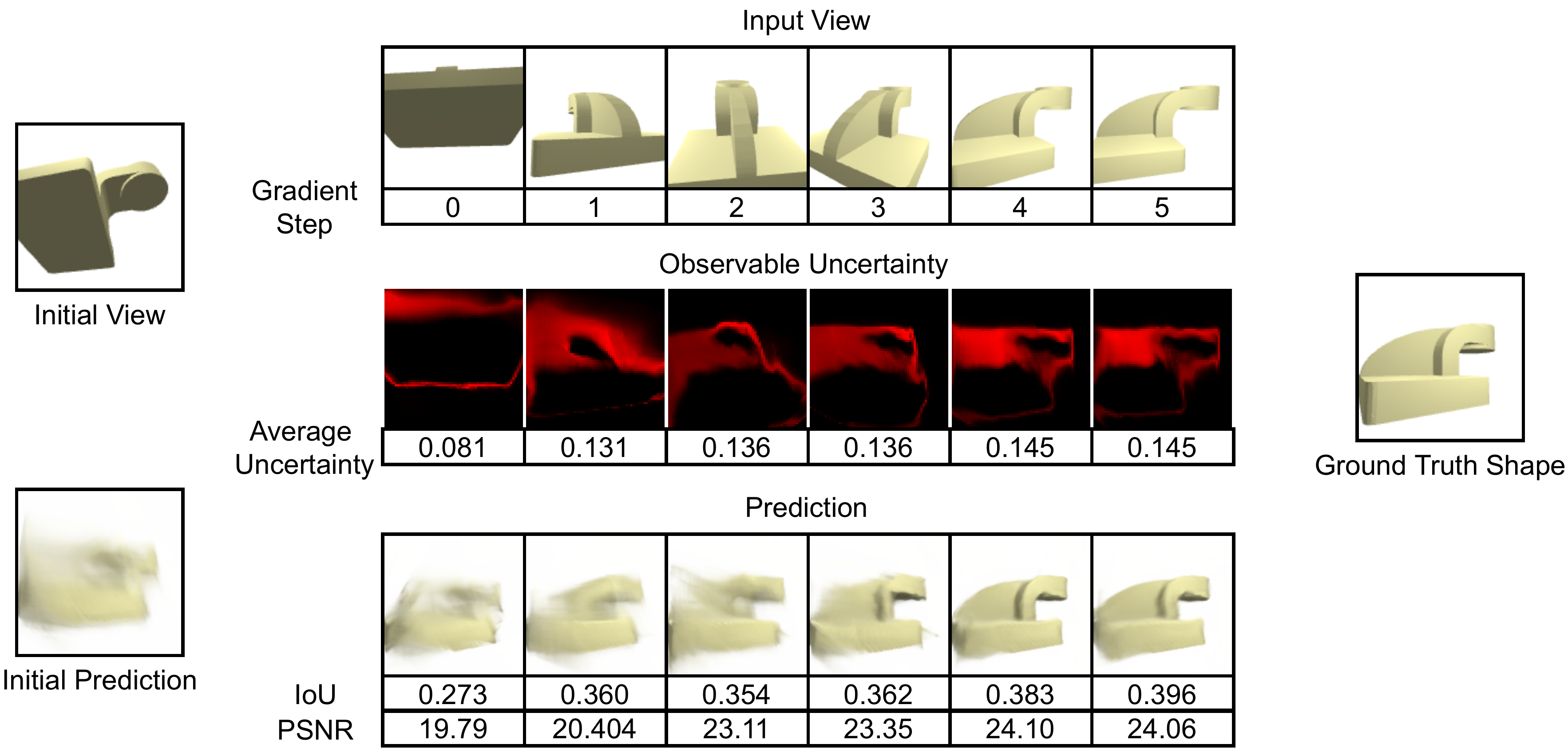}
    \caption{The change in perspective observed in the \textit{gradient} policy gradient updates in the 2D supervision setting on the ABC dataset, where for each step the current uncertainty, shape prediction, and resulting IoU and PSNR (over 20 random views) are highlighted. }
    \label{fig:2D_grad_1}
\end{figure}

\textbf{2D Supervision - ABC:}  We compare the performance of our \textit{candidate} policy with $20$ candidates to the baseline policies on the ABC dataset. Results are highlighted in Figure~\ref{fig:IoU_2D}. Additionally, we perform PSNR evaluation by sub-sampling pixel values from $20$ random views predicted using volume rendering and highlight results in Figure~\ref{fig:PSNR_2D}. Our proposed candidate policy outperforms all the baselines up to 4 selected views both in terms of IoU and PSNR. For IoU this improvement is even more pronounced than in the full 3D supervision setting. In terms of PSNR the raw improvement is small, but pronounced relative to differences among the three baselines, and worth noting as the candidate policy does not account for uncertainty in colour, and we render objects with high specularity. In the Appendix we show that over both IoU and PSNR the candidate policy significantly outperforms the baseline policies in terms of variance reduction and worse case performance. We then compare the performance of our \textit{gradient} policy to the \textit{candidate} policy and the \textit{random} policy both in terms of IoU and PSNR in Figures~\ref{fig:IoU_2D_grad} and~\ref{fig:PSNR_2D_grad} and demonstrate it effectively searches and finds areas of locally high uncertainty, leading to improved reconstruction accuracy. In Figure~\ref{fig:2D_grad_1}, we highlight the search of the gradient-based update over the perspective search space. Further candidate and gradient policy results are highlighted in the Appendix.\looseness-1

\textbf{2D Supervision - CO3D:} We further apply our \textit{candidate} policy with $10$ candidates to the baseline policies on the CO3D dataset. As volumetric ground truth shape information is not provided for this dataset, we perform this comparison in terms of PSNR only and show results in Table~\ref{table:ablation_CO3D}. We observe that our \textit{candidate} policy outperforms all baselines' selection. We note however, that while the degree to which the \textit{candidate} policy outperforms the baselines is approximately equal to the differences between them, these differences are quite small. The camera movement for CO3D image sequences tends to be on a small ring around the top of the object as opposed to the sphere of camera positions in the ABC dataset, leading far less variation in information received across images.\looseness-1 

\vspace{0.1cm}
\begin{minipage}{0.3\textwidth}
   \includegraphics[trim={5mm 3mm 3mm 3mm},clip,width=1\linewidth]{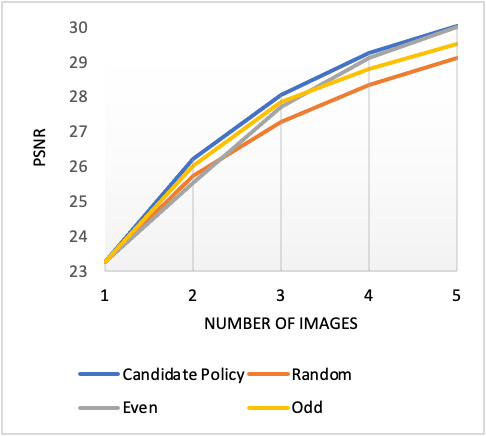}
    \captionof{figure}{PSNR comparison of policies -- 2D supervision on ABC.}
    \label{fig:PSNR_2D}
\end{minipage}\hfill
\begin{minipage}{0.3\textwidth}
    \includegraphics[trim={5mm 2mm 3mm 5.4mm},clip, width=1\linewidth]{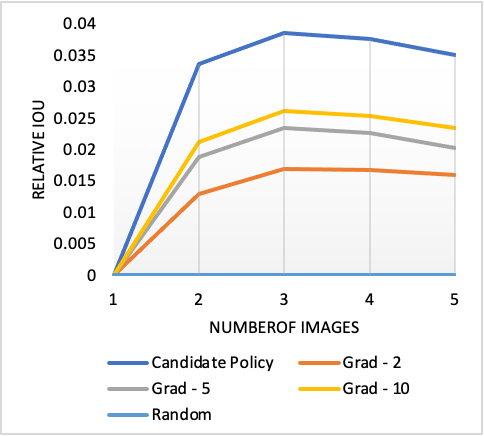}
    \captionof{figure}{IoU improvement over random policy -- 2D supervision on ABC.}
    \label{fig:IoU_2D_grad}
\end{minipage}\hfill
\begin{minipage}{0.3\textwidth}
    \includegraphics[trim={5mm 2mm 3mm 5.4mm},clip, width=1\linewidth]{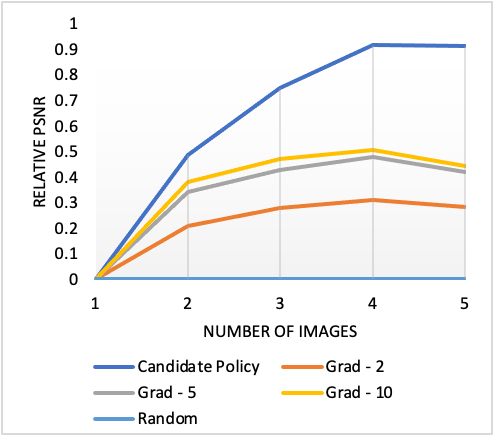}
    \captionof{figure}{PSNR improvement over random policy -- 2D supervision on ABC.}
    \label{fig:PSNR_2D_grad}
\end{minipage}
\section{Conclusion}
In this paper, we tackled next best view selection for implicit 3D scene reconstruction. We proposed an occupancy-based implicit reconstruction method which can be trained with either 3D or 2D supervision and demonstrated that it can be successfully applied to both simulated object datasets and challenging in the wild 3D scene datasets. We introduced a novel uncertainty-driven active vision method which computes the observable model uncertainty of perspectives to drive view selection, both via candidate selection and direct optimization over camera parameters. With these active view policies we demonstrated significant accuracy improvements in scene reconstruction over a set of strong baselines in both 2D and 3D supervisions settings, and over both simulated and real data, and highlighted efficient scene exploration for highly informative view.

With respect to limitations of this work, we highlight that occupancy predictions on the CO3D dataset can be quite poor, resulting in less informative uncertainty from views. In addition we also note that we currently require foreground masks of objects to operate.

\subsubsection*{Acknowledgments}
We thank Zhao Dong and Lei Xiao for their helpful feedback and comments when developing this project. We also thank David Novotny for his help in applying this work to the CO3D dataset. 

\newpage

\bibliography{iclr2023_conference}

\begin{thebibliography}{41}
\providecommand{\natexlab}[1]{#1}
\providecommand{\url}[1]{\texttt{#1}}
\expandafter\ifx\csname urlstyle\endcsname\relax
  \providecommand{\doi}[1]{doi: #1}\else
  \providecommand{\doi}{doi: \begingroup \urlstyle{rm}\Url}\fi

\bibitem[Banta et~al.(2000)Banta, Wong, Dumont, and Abidi]{banta2000next}
Joseph~E Banta, LR~Wong, Christophe Dumont, and Mongi~A Abidi.
\newblock A next-best-view system for autonomous 3-d object reconstruction.
\newblock \emph{IEEE Transactions on Systems, Man, and Cybernetics-Part A:
  Systems and Humans}, 30\penalty0 (5):\penalty0 589--598, 2000.

\bibitem[Blaer \& Allen(2007)Blaer and Allen]{blaer2007data}
Paul~S Blaer and Peter~K Allen.
\newblock Data acquisition and view planning for 3-d modeling tasks.
\newblock In \emph{2007 IEEE/RSJ International Conference on Intelligent Robots
  and Systems}, pp.\  417--422. IEEE, 2007.

\bibitem[Chen et~al.(2021)Chen, Xu, Zhao, Zhang, Xiang, Yu, and
  Su]{chen2021mvsnerf}
Anpei Chen, Zexiang Xu, Fuqiang Zhao, Xiaoshuai Zhang, Fanbo Xiang, Jingyi Yu,
  and Hao Su.
\newblock Mvsnerf: Fast generalizable radiance field reconstruction from
  multi-view stereo.
\newblock In \emph{Proceedings of the IEEE/CVF International Conference on
  Computer Vision}, pp.\  14124--14133, 2021.

\bibitem[Connolly(1985)]{connolly1985determination}
Cl~Connolly.
\newblock The determination of next best views.
\newblock In \emph{Proceedings. 1985 IEEE international conference on robotics
  and automation}, volume~2, pp.\  432--435. IEEE, 1985.

\bibitem[Fisher \& Sanchiz(1999)Fisher and Sanchiz]{fisher1999next}
Robert Fisher and Jos{\'e}~Miguel Sanchiz.
\newblock A next-best-view algorithm for 3d scene recovery with 5 degrees of
  freedom.
\newblock In \emph{Proc. British Machine Vision Conference BMVC99}, 1999.

\bibitem[Foissotte et~al.(2008)Foissotte, Stasse, Escande, and
  Kheddar]{foissotte2008next}
Torea Foissotte, Olivier Stasse, Adrien Escande, and Abderrahmane Kheddar.
\newblock A next-best-view algorithm for autonomous 3d object modeling by a
  humanoid robot.
\newblock In \emph{Humanoids 2008-8th IEEE-RAS International Conference on
  Humanoid Robots}, pp.\  333--338. IEEE, 2008.

\bibitem[Guo et~al.(2017)Guo, Pleiss, Sun, and Weinberger]{guo2017calibration}
Chuan Guo, Geoff Pleiss, Yu~Sun, and Kilian~Q Weinberger.
\newblock On calibration of modern neural networks.
\newblock In \emph{International conference on machine learning}, pp.\
  1321--1330. PMLR, 2017.

\bibitem[He et~al.(2016)He, Zhang, Ren, and Sun]{he2016deep}
Kaiming He, Xiangyu Zhang, Shaoqing Ren, and Jian Sun.
\newblock Deep residual learning for image recognition.
\newblock In \emph{Proceedings of the IEEE conference on computer vision and
  pattern recognition}, pp.\  770--778, 2016.

\bibitem[Hore \& Ziou(2010)Hore and Ziou]{hore2010image}
Alain Hore and Djemel Ziou.
\newblock Image quality metrics: Psnr vs. ssim.
\newblock In \emph{2010 20th international conference on pattern recognition},
  pp.\  2366--2369. IEEE, 2010.

\bibitem[Huber et~al.(2012)Huber, Dencker, Roschani, and
  Beyerer]{huber2012bayesian}
Marco~F Huber, Tobias Dencker, Masoud Roschani, and J{\"u}rgen Beyerer.
\newblock Bayesian active object recognition via gaussian process regression.
\newblock In \emph{2012 15th International Conference on Information Fusion},
  pp.\  1718--1725. IEEE, 2012.

\bibitem[Koch et~al.(2019)Koch, Matveev, Jiang, Williams, Artemov, Burnaev,
  Alexa, Zorin, and Panozzo]{Koch_2019_CVPR}
Sebastian Koch, Albert Matveev, Zhongshi Jiang, Francis Williams, Alexey
  Artemov, Evgeny Burnaev, Marc Alexa, Denis Zorin, and Daniele Panozzo.
\newblock Abc: A big cad model dataset for geometric deep learning.
\newblock In \emph{The IEEE Conference on Computer Vision and Pattern
  Recognition (CVPR)}, June 2019.

\bibitem[Krainin et~al.(2011)Krainin, Curless, and Fox]{krainin2011autonomous}
Michael Krainin, Brian Curless, and Dieter Fox.
\newblock Autonomous generation of complete 3d object models using next best
  view manipulation planning.
\newblock In \emph{2011 IEEE international conference on robotics and
  automation}, pp.\  5031--5037. IEEE, 2011.

\bibitem[Le et~al.(2008)Le, Saxena, and Ng]{le2008active}
Quoc~V Le, Ashutosh Saxena, and Andrew~Y Ng.
\newblock Active perception: Interactive manipulation for improving object
  detection.
\newblock \emph{Standford University Journal}, 2008.

\bibitem[Low \& Lastra(2006)Low and Lastra]{low2006adaptive}
Kok-Lim Low and Anselmo Lastra.
\newblock An adaptive hierarchical next-best-view algorithm for 3d
  reconstruction of indoor scenes.
\newblock In \emph{Proceedings of 14th Pacific Conference on Computer Graphics
  and Applications (Pacific Graphics 2006)}, pp.\  1--8, 2006.

\bibitem[Massios et~al.(1998)Massios, Fisher, et~al.]{massios1998best}
Nikolaos~A Massios, Robert~B Fisher, et~al.
\newblock \emph{A best next view selection algorithm incorporating a quality
  criterion}, volume~2.
\newblock Department of Artificial Intelligence, University of Edinburgh, 1998.

\bibitem[Mendoza et~al.(2020)Mendoza, Vasquez-Gomez, Taud, Sucar, and
  Reta]{mendoza2020supervised}
Miguel Mendoza, J~Irving Vasquez-Gomez, Hind Taud, L~Enrique Sucar, and
  Carolina Reta.
\newblock Supervised learning of the next-best-view for 3d object
  reconstruction.
\newblock \emph{Pattern Recognition Letters}, 133:\penalty0 224--231, 2020.

\bibitem[Mescheder et~al.(2019)Mescheder, Oechsle, Niemeyer, Nowozin, and
  Geiger]{mescheder2019occupancy}
Lars Mescheder, Michael Oechsle, Michael Niemeyer, Sebastian Nowozin, and
  Andreas Geiger.
\newblock Occupancy networks: Learning 3d reconstruction in function space.
\newblock In \emph{Proceedings of the IEEE/CVF conference on computer vision
  and pattern recognition}, pp.\  4460--4470, 2019.

\bibitem[Mildenhall et~al.(2020)Mildenhall, Srinivasan, Tancik, Barron,
  Ramamoorthi, and Ng]{mildenhall2020nerf}
Ben Mildenhall, Pratul~P Srinivasan, Matthew Tancik, Jonathan~T Barron, Ravi
  Ramamoorthi, and Ren Ng.
\newblock Nerf: Representing scenes as neural radiance fields for view
  synthesis.
\newblock In \emph{European conference on computer vision}, pp.\  405--421.
  Springer, 2020.

\bibitem[Oechsle et~al.(2021)Oechsle, Peng, and Geiger]{oechsle2021unisurf}
Michael Oechsle, Songyou Peng, and Andreas Geiger.
\newblock Unisurf: Unifying neural implicit surfaces and radiance fields for
  multi-view reconstruction.
\newblock In \emph{Proceedings of the IEEE/CVF International Conference on
  Computer Vision}, pp.\  5589--5599, 2021.

\bibitem[Park et~al.(2019)Park, Florence, Straub, Newcombe, and
  Lovegrove]{park2019deepsdf}
Jeong~Joon Park, Peter Florence, Julian Straub, Richard Newcombe, and Steven
  Lovegrove.
\newblock Deepsdf: Learning continuous signed distance functions for shape
  representation.
\newblock In \emph{Proceedings of the IEEE/CVF conference on computer vision
  and pattern recognition}, pp.\  165--174, 2019.

\bibitem[Peng et~al.(2020)Peng, Wang, Miao, Feng, and Tang]{peng2020viewpoints}
Weixing Peng, Yaonan Wang, Zhiqiang Miao, Mingtao Feng, and Yongpeng Tang.
\newblock Viewpoints planning for active 3-d reconstruction of profiled blades
  using estimated occupancy probabilities (eop).
\newblock \emph{IEEE Transactions on Industrial Electronics}, 68\penalty0
  (5):\penalty0 4109--4119, 2020.

\bibitem[Peralta et~al.(2020)Peralta, Casimiro, Nilles, Aguilar, Atienza, and
  Cajote]{peralta2020next}
Daryl Peralta, Joel Casimiro, Aldrin~Michael Nilles, Justine~Aletta Aguilar,
  Rowel Atienza, and Rhandley Cajote.
\newblock Next-best view policy for 3d reconstruction.
\newblock In \emph{European Conference on Computer Vision}, pp.\  558--573.
  Springer, 2020.

\bibitem[Pito(1999)]{pito1999solution}
Richard Pito.
\newblock A solution to the next best view problem for automated surface
  acquisition.
\newblock \emph{IEEE Transactions on pattern analysis and machine
  intelligence}, 21\penalty0 (10):\penalty0 1016--1030, 1999.

\bibitem[Reizenstein et~al.(2021)Reizenstein, Shapovalov, Henzler, Sbordone,
  Labatut, and Novotny]{reizenstein2021common}
Jeremy Reizenstein, Roman Shapovalov, Philipp Henzler, Luca Sbordone, Patrick
  Labatut, and David Novotny.
\newblock Common objects in 3d: Large-scale learning and evaluation of
  real-life 3d category reconstruction.
\newblock In \emph{Proceedings of the IEEE/CVF International Conference on
  Computer Vision}, pp.\  10901--10911, 2021.

\bibitem[Schwarz et~al.(2020)Schwarz, Liao, Niemeyer, and
  Geiger]{schwarz2020graf}
Katja Schwarz, Yiyi Liao, Michael Niemeyer, and Andreas Geiger.
\newblock Graf: Generative radiance fields for 3d-aware image synthesis.
\newblock \emph{Advances in Neural Information Processing Systems},
  33:\penalty0 20154--20166, 2020.

\bibitem[Sebastian et~al.(2005)Sebastian, Wolfram, and
  Dieter]{sebastian2005probabilistic}
THRUN Sebastian, Burgard Wolfram, and F~Dieter.
\newblock Probabilistic robotics, 2005.

\bibitem[Simonyan \& Zisserman(2014)Simonyan and Zisserman]{simonyan2014very}
Karen Simonyan and Andrew Zisserman.
\newblock Very deep convolutional networks for large-scale image recognition.
\newblock \emph{arXiv preprint arXiv:1409.1556}, 2014.

\bibitem[Smith et~al.(2021)Smith, Meger, Pineda, Calandra, Malik,
  Romero~Soriano, and Drozdzal]{smith2021active}
Edward Smith, David Meger, Luis Pineda, Roberto Calandra, Jitendra Malik,
  Adriana Romero~Soriano, and Michal Drozdzal.
\newblock Active 3d shape reconstruction from vision and touch.
\newblock \emph{Advances in Neural Information Processing Systems},
  34:\penalty0 16064--16078, 2021.

\bibitem[Vasquez-Gomez et~al.(2013)Vasquez-Gomez, Sucar, and
  Murrieta-Cid]{vasquez2013hierarchical}
J~Irving Vasquez-Gomez, L~Enrique Sucar, and Rafael Murrieta-Cid.
\newblock Hierarchical ray tracing for fast volumetric next-best-view planning.
\newblock In \emph{2013 International Conference on Computer and Robot Vision},
  pp.\  181--187. IEEE, 2013.

\bibitem[Vasquez-Gomez et~al.(2014)Vasquez-Gomez, Sucar, Murrieta-Cid, and
  Lopez-Damian]{vasquez2014volumetric}
J~Irving Vasquez-Gomez, L~Enrique Sucar, Rafael Murrieta-Cid, and Efrain
  Lopez-Damian.
\newblock Volumetric next-best-view planning for 3d object reconstruction with
  positioning error.
\newblock \emph{International Journal of Advanced Robotic Systems}, 11\penalty0
  (10):\penalty0 159, 2014.

\bibitem[Vasquez-Gomez et~al.(2021)Vasquez-Gomez, Troncoso, Becerra, Sucar, and
  Murrieta-Cid]{vasquez2021next}
J~Irving Vasquez-Gomez, David Troncoso, Israel Becerra, Enrique Sucar, and
  Rafael Murrieta-Cid.
\newblock Next-best-view regression using a 3d convolutional neural network.
\newblock \emph{Machine Vision and Applications}, 32\penalty0 (2):\penalty0
  1--14, 2021.

\bibitem[Wang et~al.(2018)Wang, Zhang, Li, Fu, Liu, and
  Jiang]{wang2018pixel2mesh}
Nanyang Wang, Yinda Zhang, Zhuwen Li, Yanwei Fu, Wei Liu, and Yu-Gang Jiang.
\newblock Pixel2mesh: Generating 3d mesh models from single rgb images.
\newblock In \emph{Proceedings of the European conference on computer vision
  (ECCV)}, pp.\  52--67, 2018.

\bibitem[Watt et~al.(2020)Watt, Borhani, and K.~Katsaggelos]{mlbook}
Jeremy Watt, Reza Borhani, and Aggelos K.~Katsaggelos.
\newblock \emph{Machine Learning Refined: Foundations, Algorithms, and
  Applications}.
\newblock Cambridge University Press;, Cambridge, 2020.

\bibitem[Wu et~al.(2018)Wu, Wang, Yin, Lu, and Zhu]{wu2018sharing}
Huijun Wu, Chen Wang, Jie Yin, Kai Lu, and Liming Zhu.
\newblock Sharing deep neural network models with interpretation.
\newblock In \emph{Proceedings of the 2018 World Wide Web Conference}, pp.\
  177--186, 2018.

\bibitem[Xie et~al.(2019)Xie, Yao, Sun, Zhou, and Zhang]{xie2019pix2vox}
Haozhe Xie, Hongxun Yao, Xiaoshuai Sun, Shangchen Zhou, and Shengping Zhang.
\newblock Pix2vox: Context-aware 3d reconstruction from single and multi-view
  images.
\newblock In \emph{Proceedings of the IEEE/CVF international conference on
  computer vision}, pp.\  2690--2698, 2019.

\bibitem[Yang et~al.(2018)Yang, Wang, Wang, Yin, Zhang, Wei, and
  Fu]{yang2018active}
Xin Yang, Yuanbo Wang, Yaru Wang, Baocai Yin, Qiang Zhang, Xiaopeng Wei, and
  Hongbo Fu.
\newblock Active object reconstruction using a guided view planner.
\newblock \emph{arXiv preprint arXiv:1805.03081}, 2018.

\bibitem[Yariv et~al.(2020)Yariv, Kasten, Moran, Galun, Atzmon, Ronen, and
  Lipman]{yariv2020multiview}
Lior Yariv, Yoni Kasten, Dror Moran, Meirav Galun, Matan Atzmon, Basri Ronen,
  and Yaron Lipman.
\newblock Multiview neural surface reconstruction by disentangling geometry and
  appearance.
\newblock \emph{Advances in Neural Information Processing Systems},
  33:\penalty0 2492--2502, 2020.

\bibitem[Yu et~al.(2021)Yu, Ye, Tancik, and Kanazawa]{yu2021pixelnerf}
Alex Yu, Vickie Ye, Matthew Tancik, and Angjoo Kanazawa.
\newblock pixelnerf: Neural radiance fields from one or few images.
\newblock In \emph{Proceedings of the IEEE/CVF Conference on Computer Vision
  and Pattern Recognition}, pp.\  4578--4587, 2021.

\bibitem[Zaheer et~al.(2017)Zaheer, Kottur, Ravanbakhsh, Poczos, Salakhutdinov,
  and Smola]{zaheer2017deep}
Manzil Zaheer, Satwik Kottur, Siamak Ravanbakhsh, Barnabas Poczos, Russ~R
  Salakhutdinov, and Alexander~J Smola.
\newblock Deep sets.
\newblock \emph{Advances in neural information processing systems}, 30, 2017.

\bibitem[Zhou et~al.(2019)Zhou, Fang, Song, Guan, Yin, Dai, and
  Yang]{zhou2019iou}
Dingfu Zhou, Jin Fang, Xibin Song, Chenye Guan, Junbo Yin, Yuchao Dai, and
  Ruigang Yang.
\newblock Iou loss for 2d/3d object detection.
\newblock In \emph{2019 International Conference on 3D Vision (3DV)}, pp.\
  85--94. IEEE, 2019.

\bibitem[Zhou et~al.(2012)Zhou, Kantarcioglu, and Thuraisingham]{zhou2012self}
Yan Zhou, Murat Kantarcioglu, and Bhavani Thuraisingham.
\newblock Self-training with selection-by-rejection.
\newblock In \emph{2012 IEEE 12th international conference on data mining},
  pp.\  795--803. IEEE, 2012.

\end{thebibliography}
\bibliographystyle{iclr2023_conference}

\appendix
\section{Appendix}

\subsection{Reconstruction Model Details }
The reconstruction model used to predict 3D scene properties in the 3D supervision setting takes as input a set of $K$ images, camera positions, and camera orientations in quaternion format, and a position in space. The set of $K$ images are passed independently through a large CNN. The architecture of this CNN is highlighted in Table~\ref{table:3D_Network}, and features from layers 7, 13, and 19 and 25 are collected and sampled using perceptual image pooling~\citep{wang2018pixel2mesh} leading to a feature vector of size 480 which is passed through two fully connected layers to change its size to 125. The position information and camera information from each camera are all passed through positional embedding layers~\citep{mildenhall2020nerf}, concatenated, and then passed through a single fully connected layer leading to a feature vector of size 125. The image features and positions features are then concatenated and passed through a set of 3 ResNet blocks~\citep{he2016deep} with hidden size 250. This leads to $K$ features vectors of size 250. To aggregate the information from multiple cameras, we use a set of 3 deep set layers with max pooling and hidden size 250~\citep{zaheer2017deep}, and then apply a final fully connected layer with sigmoid activation to predict the occupancy value. \looseness-1

The reconstruction model used to predict 3D scene properties in the 2D supervision setting on the ABC dataset takes as input a set of $K$ images, camera positions, and camera orientations in quaternion format, a position in space, and a target camera's position and orientation in quaternion format. The set of $K$ images are passed independently through a large CNN. The architecture of this CNN is highlighted in Table~\ref{table:2D_Network}, and features from layers 7, 13, and 19 and 25 are collected and sampled using perceptual image pooling~\citep{wang2018pixel2mesh} leading to a feature vector of size 600 which is passed through a two fully connected layers to change its size to 125. The position information and camera information from each camera are all passed through positional embedding layers~\citep{mildenhall2020nerf}, concatenated, and then passed through a single fully connected layer leading to a feature vector of size 125. The image features and positions features are then concatenated and passed through a set of 3 ResNet blocks~\citep{he2016deep} with hidden size 250. This leads to $K$ features vectors of size 250. To aggregate the information from multiple cameras, we use a set of 3 deep set layers with max pooling and hidden size 250, and then apply a final fully connected layer with sigmoid activation to predict the occupancy value. Intermediary features from this aggregation and a positional embedding of the target camera's parameters are then concatenated and passed through 3 fully connected layers with a final sigmoid activation to predict the colour. For training on the CO3D dataset, the operations are identical except that the input images are of size $256\times256$ and so the height and width of image feature maps are doubled. 

All training is performed on a single Tesla V100 GPU with 16 CPU cores with the Adam optimizer and learning rate $1e^{-4}$ for 3 days, models selected using grid search over networks lengths and layer widths, and training stopped frozen using early stopping.

\subsection{Reconstruction Results}
\begin{figure}[H]
    \centering
    \includegraphics[width=\linewidth]{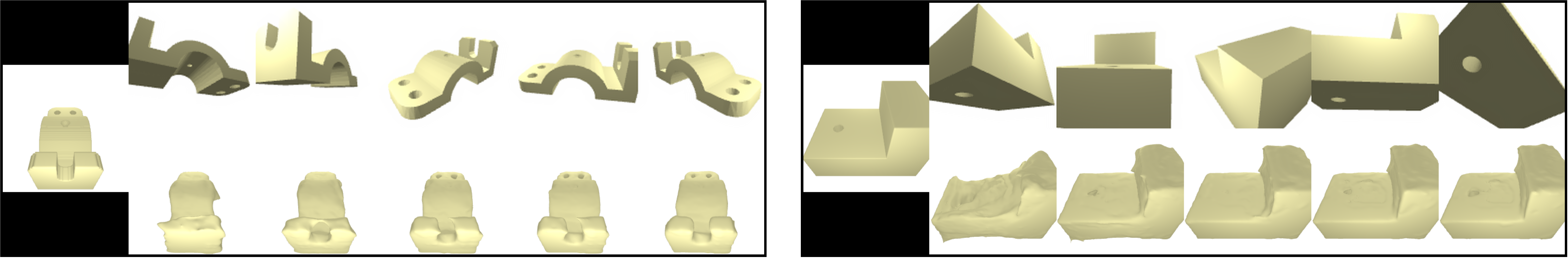}
    \caption{Reconstruction results on the ABC dataset in the 3D supervision setting where in each example the left most image is the ground truth shape, the upper row is the input images, and the lower row is the predicted shape with 1 to 5 images leveraged. }
    \label{fig:3D_recon_sup}
\end{figure}

\begin{figure}[H]
    \centering
    \includegraphics[width=\linewidth]{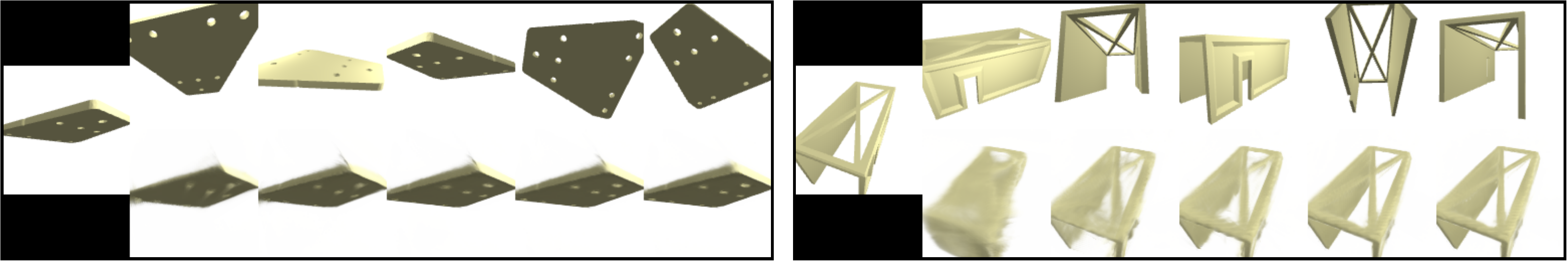}
    \caption{Reconstruction results on the ABC dataset in the 2D supervision setting where in each example the left most image is the ground truth shape, the upper row is the input images, and the lower row is the rendered shape with 1 to 5 images leveraged. }
    \label{fig:2D_recon_sup}
\end{figure}

We visualize predicted shape examples for the 3D setting on the ABC dataset in Figure~\ref{fig:3D_recon_sup} and rendered shape examples for the 2D setting on the ABC dataset in Figure~\ref{fig:2D_recon_sup}. We provide PSNR results on the test set with 2D supervision training on the CO3D and ABC datasets in Table~\ref{table:psnr_2d_results}.

\subsection{Uncertainty Analysis}
\begin{figure}[t]
    \centering
    \includegraphics[width=0.8\linewidth]{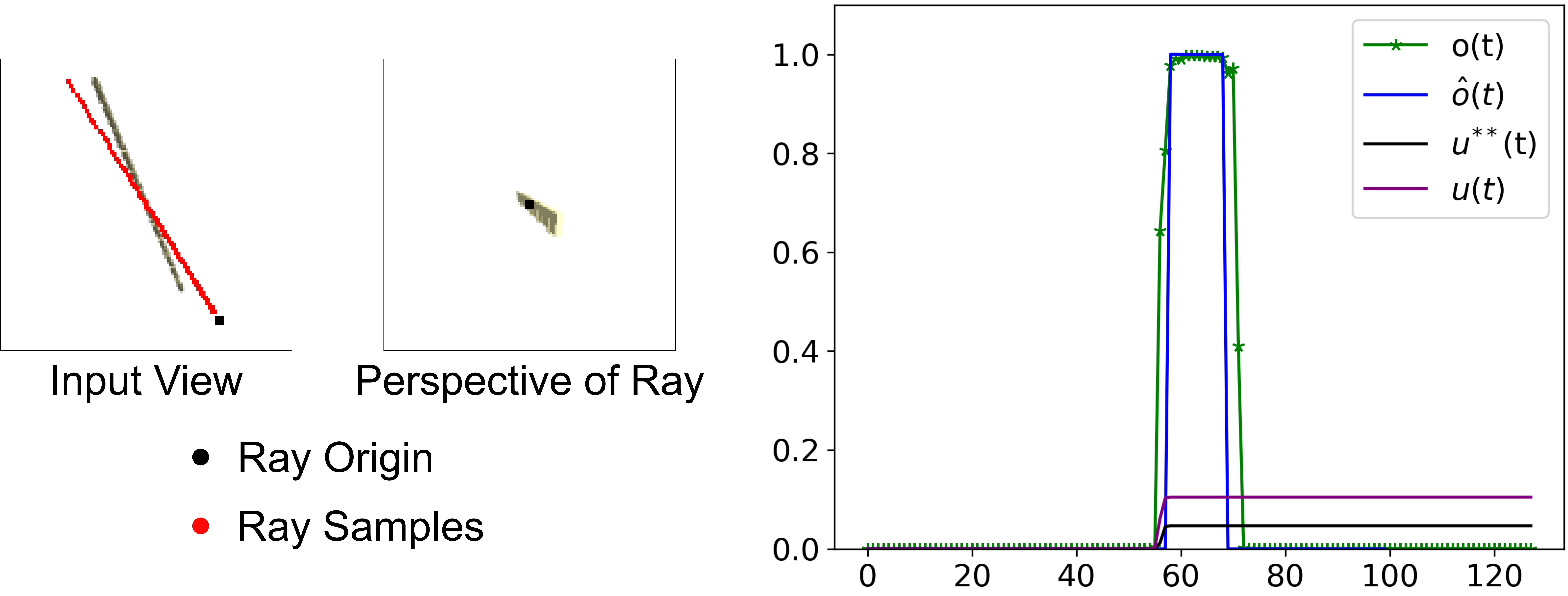}
    \caption{Demonstration of the accumulation of uncertainty along a ray. On the left we display the initial input image for to the model and in the middle a new perspective into the scene for a ray, with the ray origin labeled in black, and samples along the ray labeled in red. In the graph on the left we highlight the per-sample ground truth occupancy values $\hat{o}(t)$, predicted occupancy values $o(t)$, and the resulting accumulated uncertainty both with the rate of change correction, $u(t)$, and without it $u^{**}(t)$. }
    \label{fig:ray_dir}
\end{figure}

We provide further analysis of our uncertainty definition along rays. In Figure \ref{fig:ray_dir} we highlight how predicted occupancy values along a ray through a scene results in the accumulation of uncertainty both with and without the use of the directional derivative. In this example the model is very confident about the occupancy of the object at every point along the ray. Due to its limited resolution, however, some occupancy's are predicted to be close to the decision boundary and, without the presence of the rate of change correction, this results in the accumulation of significant uncertainty.

\begin{figure}[t]
    \centering
    \includegraphics[width=0.8\linewidth]{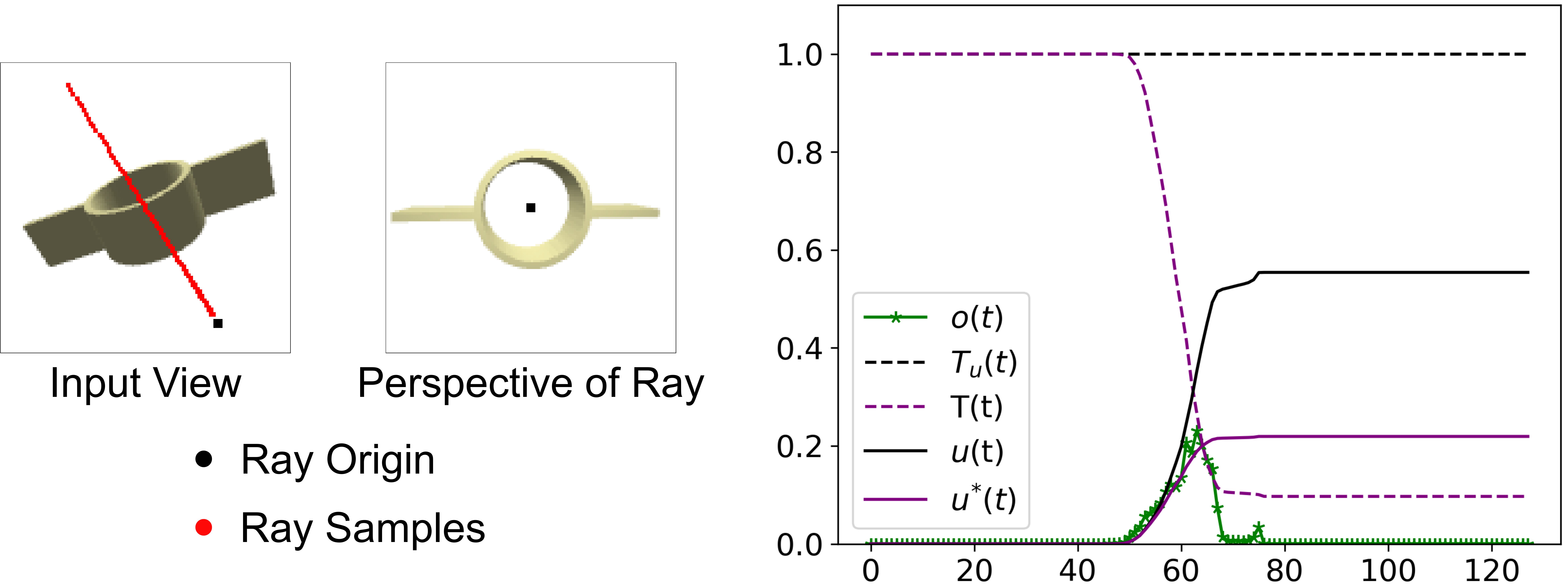}
    \caption{Demonstration of the accumulation of uncertainty along a ray. On the left we display the initial input image for to the model and in the middle a new perspective into the scene for a ray, with the ray origin labeled in black, and samples along the ray labeled in red. In the graph on the left we highlight the per-sample predicted occupancy values $o(t)$, and resulting uncertainty accumulation function values $T_u(t)$ and accumulated uncertainty $u(t)$. In addition we also provide resulting uncertainty accumulation function values $T(t)$ and accumulated uncertainty $u^{*}(t)$ if the standard volume rending accumulation function is leveraged (equation \ref{eq:VR}).}
    \label{fig:ray_acc}
\end{figure}

In Figure \ref{fig:ray_acc} we highlight how predicted occupancy values along a ray through a scene results in the accumulation of uncertainty both when using the original volume rendering accumulation function, $T(t)$, and when using our uncertainty based accumulation function, $T_u(t)$. In this example the model is uncertain if the ray intersects with the object's surface, as indicated by predicted occupancy values hovering around $0.3$ for a short period. With the standard volume rendering accumulation (equation \ref{eq:VR}) any predicted occupancy results in the reduction of accumulation for future ray samples as shown by the reduction of $T(t)$ from $1$ to around $0.2$. We avoid this reduction by leveraging our uncertainty based accumulation function which does not penalize negative occupancy predictions ($o(t) < 0.5$). As a result the accumulated uncertainty which leverages $T(t)$,  $u^{*}(t)$, is significantly lower then our true accumulated uncertainty, $u(t)$, by the end of the ray.

\subsection{Next Best View Details}
In the 3D supervision setting for the ABC Dataset, we set $\lambda_s = 2$, $\lambda_u = 2$, $\lambda_t = 4$, $\lambda_d = .5$,  $\lambda_D = .05$, $\lambda_S = 4$ and $\lambda = 1$. The minimum distance of views in the \textit{candidate} selection policy is $0.7$. For the \textit{gradient} selection policy, the learning rate is set to $0.5$ and optimization is performed using the Adam optimizer. 

In the 2D supervision setting for the ABC Dataset, we set $\lambda_s = 0.5$, $\lambda_u = 4$ , $\lambda_t = 4$, $\lambda_d = [10, 2, .5, .2]$ for selection 1 to 4, $\lambda_D = .05$, $\lambda_S = 4$, and $\lambda = 0$. The minimum distance of views in the \textit{candidate} selection policy is $0.7$. For the \textit{gradient} selection policy, the learning rate is set to $0.5$ and optimization is performed using the Adam optimizer. In the 2D supervision setting for the CO3D dataset, we leverage only the silhouette uncertainty with $\lambda_s = 0.5$ as this model's local occupancy predictions tended to be poor due, and this setting performed better empirically better on the validation set.

\subsection{Next Best View Ablations}
Here we provide an ablation over choices made in our proposed next best view method. This ablation in the 3D supervision setting is highlighted in Table~\ref{table:ablation_3D} where we highlight that removing the silhouette uncertainty, depth uncertainty, uncertainty based accumulation function, and rate of change correction leads to reduction in IoU accuracy. We do note that for the first action selection, removing the rate of change slightly improves accuracy, but that for all further action selections its addition significantly improves performance. The Ablation in the 2D supervision setting is highlighted in Tables~\ref{table:ablation_2D_iou} and~\ref{table:ablation_2D_psnr} for IoU and PSNR. We highlight that removing the silhouette uncertainty, depth uncertainty, uncertainty based accumulation function, and rate of change correction leads to reduction in IoU accuracy. For PSNR, the same is observed except that the removal of the rate uncertainty based accumulation function, and rate of change correction leads to slightly worse PSNR results.\looseness-1 

\begin{figure}[h]
\begin{floatrow}
\capbtabbox{
\scalebox{1}{
  \vspace{0.2cm}
  \scalebox{0.6}{
 \begin{tabular}[H]{cccccc}
    \toprule

    \multicolumn{1}{c}{} & \multicolumn{5}{c}{\textbf{Number of Images}}  \\
    
     \textbf{Model Setting} & 1 & 2 & 3 & 4 & 5 \\
    
    \cmidrule[1pt](lr){1-1}\cmidrule[1pt](lr){2-6}
   
     Full Policy & 0.6218	& 0.7177	& \textbf{0.7565}	& \textbf{0.7755}	& \textbf{0.7861} \\
     No Sil. & 0.6218 &	0.7149 &	0.7551 &	0.7745 &	0.7854 \\
     No Depth &  0.6218	& 0.7041 &	0.7420	& 0.7649	& 0.7789 \\
     No $T_u(t)$ & 0.6218 &	0.7177 &	0.7536 &	0.7719 &	0.7829 \\
     No $d(t)$ & 0.6218	 & \textbf{0.7185} & 	0.7541 & 	0.7727 & 	0.7836 \\

    \cmidrule[1pt](lr){1-1}\cmidrule[1pt](lr){2-6} \\

    \bottomrule
 \end{tabular}
 }}}
 {%
   \centering
  \caption{Ablation over next best view performance in the 3D supervision setting on the ABC dataset.}
  \label{table:ablation_3D}
  \small
 }
\capbtabbox{
\scalebox{1}{
  \vspace{0.2cm}
  \scalebox{0.6}{
  \begin{tabular}{cccccc}
    \toprule
    \multicolumn{1}{c}{} & \multicolumn{5}{c}{\textbf{Number of Images}}  \\
     \textbf{Dataset} & 1 & 2 & 3 & 4 & 5 \\
    \cmidrule[1pt](lr){1-1}\cmidrule[1pt](lr){2-6}
   ABC & 21.063 & 23.238 & 24.708 & 25.890 & 27.164 \\
    \cmidrule[1pt](lr){1-1}\cmidrule[1pt](lr){2-6}
    Co3D & 16.776 & 17.645 & 18.234 & 18.750 &  18.797 \\
    \bottomrule
 \end{tabular}
 }}}
 {%
  \centering
  \caption{PSNR reconstruction results on the test sets from training on the ABC and CO3D daatasets }
  \label{table:psnr_2d_results}
  \small
 }

\end{floatrow}
\end{figure}

\begin{figure}[h]
\begin{floatrow}
\capbtabbox{
\scalebox{1}{
  \vspace{0.2cm}
  \scalebox{0.6}{
  \begin{tabular}[H]{cccccc}
    \toprule

    \multicolumn{1}{c}{} & \multicolumn{5}{c}{\textbf{Number of Images}}  \\
    
     \textbf{Model Setting} & 1 & 2 & 3 & 4 & 5 \\
    
    \cmidrule[1pt](lr){1-1}\cmidrule[1pt](lr){2-6}
   
     Full Policy & 0.2996 &	\textbf{0.4437} &\textbf{0.5062} &	\textbf{0.5419}	& \textbf{0.5644} \\
     No Sil. & 0.2996&	0.4365	& 0.5045	& 0.5410	 & \textbf{0.5644} \\
     No Depth & 0.2996	& 0.4256 & 	0.4882	& 0.5292	& 0.5561\\
     No $T_u(t)$ & 0.2996	& 0.4430	 & 0.5047	 & 0.5403	& 0.5627 \\
     No $d(t)$ & 0.2996  & 	0.4436	& 0.5060	 & 0.5416	& 0.5629 \\

    \cmidrule[1pt](lr){1-1}\cmidrule[1pt](lr){2-6} \\

    \bottomrule
 \end{tabular}
 }}}
 {%
  \centering
  \caption{Ablation results for IoU on the ABC dataset in the 2D supervision setting.}
  \label{table:ablation_2D_iou}
  \small
 }
\capbtabbox{
\scalebox{1}{
  \vspace{0.2cm}
  \scalebox{0.6}{
  \begin{tabular}[H]{cccccc}
    \toprule

    \multicolumn{1}{c}{} & \multicolumn{5}{c}{\textbf{Number of Images}}  \\
    
     \textbf{Model Setting} & 1 & 2 & 3 & 4 & 5 \\
    
    \cmidrule[1pt](lr){1-1}\cmidrule[1pt](lr){2-6}
   
      Full Policy & 23.2532	& 26.2307	& 28.0478	& 29.2633	& 30.0405 \\
     No Sil. & 23.2532	& 26.0313	& 27.7125	& 28.8475	& 29.7231\\
     No Depth & 23.2532	 & 26.1232	& 27.8261	& 29.0386	& 29.9100\\
     No $T_u(t)$ &23.2532	& \textbf{26.2615}	 & \textbf{28.0748}	 & \textbf{29.2888}	 & 30.0762\\
     No $d(t)$ & 23.2532	& 26.2316	& 28.0555	& 29.2836	& \textbf{30.0812} \\

    \cmidrule[1pt](lr){1-1}\cmidrule[1pt](lr){2-6} \\

    \bottomrule
 \end{tabular}
 }}}
 {%
  \centering
  \caption{Ablation results for PSNR on the ABC dataset in the 2D supervision setting.}
  \label{table:ablation_2D_psnr}
  \small
 }

\end{floatrow}
\end{figure}

\subsection{Next Best View Additional Results}
\begin{figure}[H]
    \centering
    \includegraphics[width=.8\linewidth]{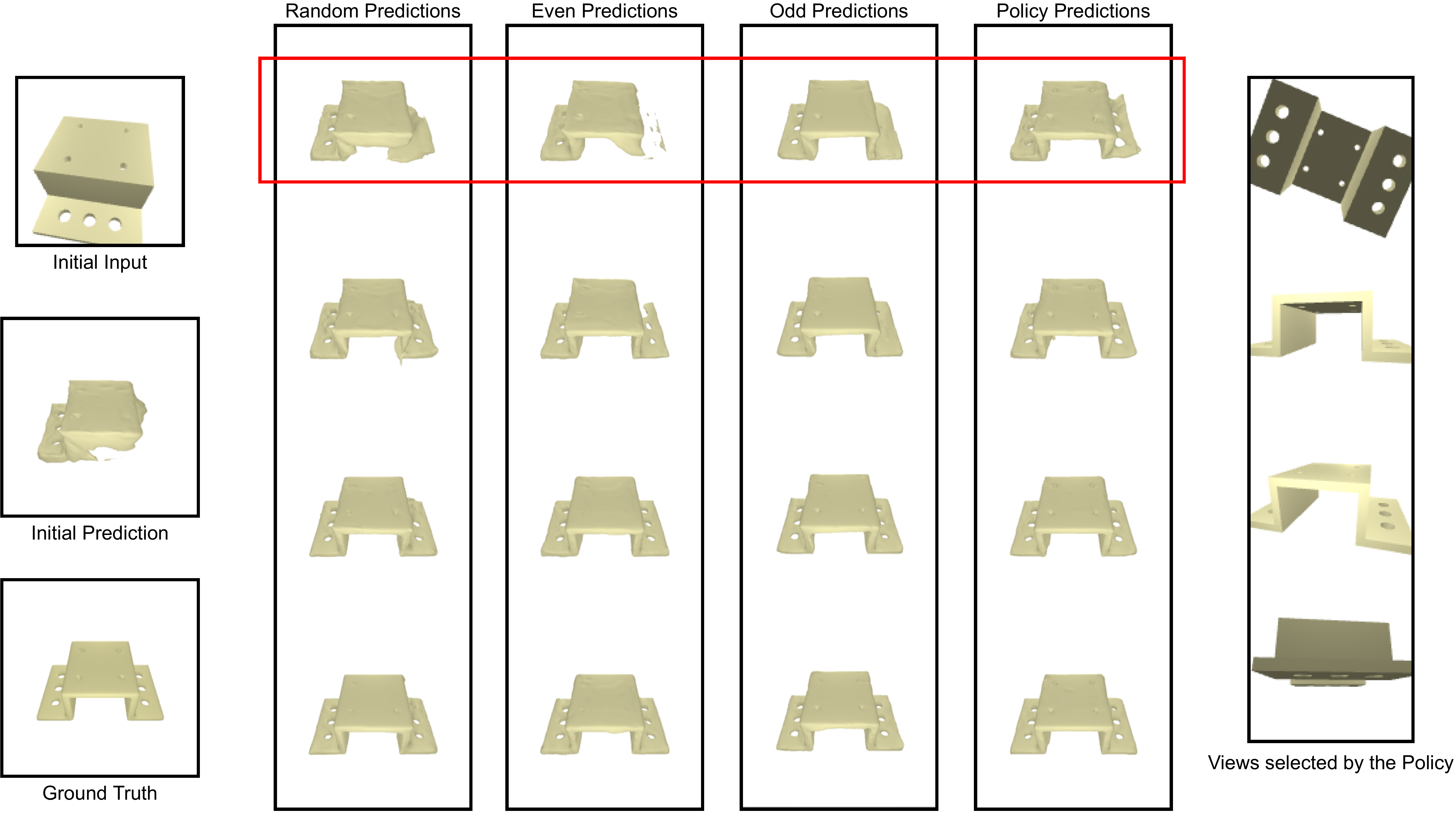}
    \caption{Shape reconstruction comparison over selection policies, with the \textit{candidate} policy view selections highlighted in final column. Results shown for ABC in the 3D supervision setting.}
    \label{fig:3D_comparision_2}
\end{figure}

\begin{figure}[H]
    \centering
    \includegraphics[width=.8\linewidth]{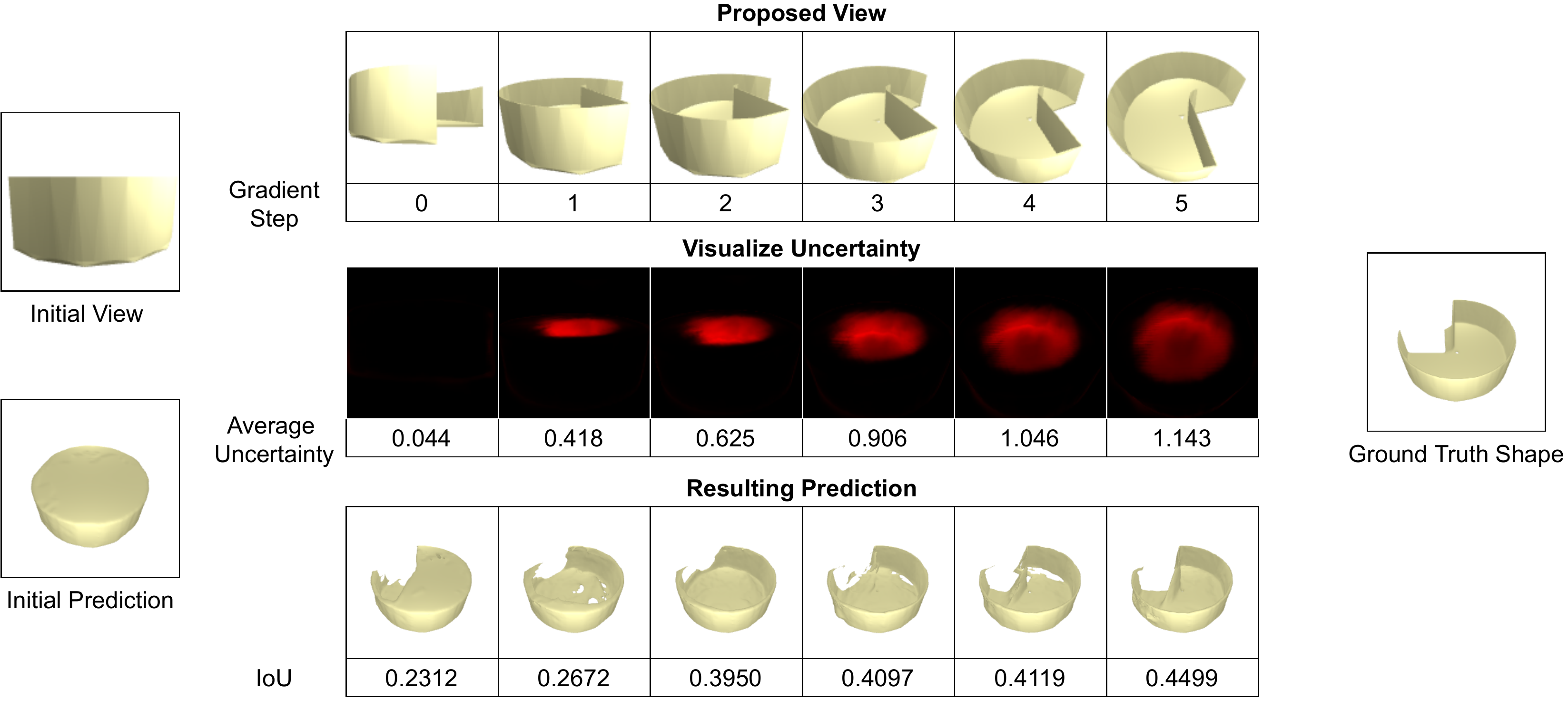}
    \caption{The change in perspective observed in the \textit{gradient} policy gradient updates, where for each step the current uncertainty, shape prediction, and resulting IoU are highlighted. Results shown for ABC in the 3D supervision setting.\looseness-1}
    \label{fig:3D_grad_1}
\end{figure}

\begin{figure}[H]
    \centering
    \includegraphics[width=.8\linewidth]{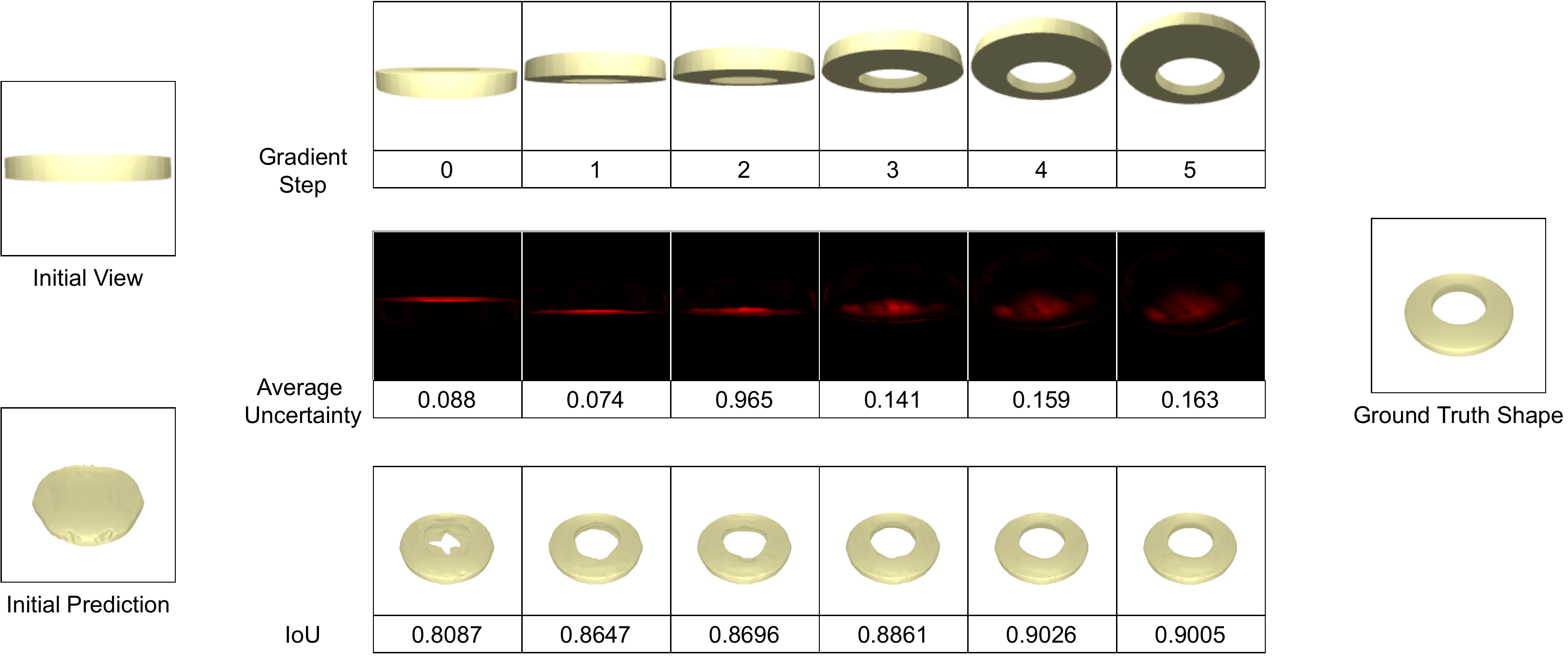}
    \caption{The change in perspective observed in the \textit{gradient} policy gradient updates, where for each step the current uncertainty, shape prediction, and resulting IoU are highlighted. Results shown on ABC in the 3D supervision setting.}
    \label{fig:3D_grad_2}
\end{figure}

We highlight the performance of all policies over IoU in the 3D supervision setting on the ABC dataset in Table~\ref{table:policy_recon_IoU}. This evaluation is with 10 random view initializations and over the 500 test set objects. We demonstrate in Table~\ref{table:policy_recon_IoU_worst} the worst case performance of all policies over the 10 random initializations, and show that our \textit{candidate} policy continues to our outperform all others. We demonstrate in Table~\ref{table:policy_recon_IoU_STD}  standard deviation in IoU of all policies over the 10 random initializations, and show that our \textit{candidate} policy also results in lower variance of performance. In Figure~\ref{fig:3D_comparision_2}, we provide additional reconstruction results for different policy selection strategies and in Figures~\ref{fig:3D_grad_1} and~\ref{fig:3D_grad_2}, we provide additional results for the \textit{gradient} policy's updates over the perspective search space.\looseness-1

\begin{table*}
  \centering
  \caption{IoU performance of all policies on the ABC test set in the 3D supervision setting.}
  \label{table:policy_recon_IoU}
  \small
\begin{tabular}[H]{cccccc}
    \toprule

    \multicolumn{1}{c}{} & \multicolumn{5}{c}{\textbf{Number of Images}}  \\
    
     \textbf{Model Setting} & 1 & 2 & 3 & 4 & 5 \\
    
    \cmidrule[1pt](lr){1-1}\cmidrule[1pt](lr){2-6}
   
     Candidate Policy &  0.6218	& \textbf{0.7177}	& \textbf{0.7565}	& \textbf{0.7755}	& \textbf{0.7861}\\
     Grad - 2 Steps & 0.6218 & 0.7082& 0.7459& 0.7657& 0.7778 \\
     Grad - 5 Steps & 0.6218 & 0.7123 &0.7487 &0.7665 &0.7772\\
     Grad - 10 Steps & 0.6218 &0.7135 &0.7491 &0.7663 &0.7773\\
     Random & 0.6218 & 0.7044 & 0.7424& 0.7629& 0.7756\\
     Even & 0.6218 & 0.6945& 0.7418& 0.7669& 0.7798\\
     Odd & 0.6218& 0.7068& 0.7464& 0.7620 & 0.7730\\
    \bottomrule
 \end{tabular}%}
\end{table*}

\begin{table*}
  \centering
  \caption{Worst case IoU performance of all policies on the ABC test set in the 3D supervision setting.\looseness-1}
  \label{table:policy_recon_IoU_worst}
  \small

\begin{tabular}[H]{cccccc}
    \toprule

    \multicolumn{1}{c}{} & \multicolumn{5}{c}{\textbf{Number of Images}}  \\
    
     \textbf{Model Setting} & 1 & 2 & 3 & 4 & 5 \\
    
    \cmidrule[1pt](lr){1-1}\cmidrule[1pt](lr){2-6}
   
     Candidate Policy & 0.4696 & \textbf{0.6335} & \textbf{0.6959} & \textbf{0.7298} &\textbf{0.7474} \\
     Grad - 2 Steps  & 0.4696 &0.6128 &0.6714 &0.7066 &0.7285 \\
     Grad - 5 Steps &  0.4696 &0.6205 &0.6784 &0.7107 &0.7281\\
     Grad - 10 Steps & 0.4696 &0.6193 &0.6775 &0.7085 &0.7259\\
     Random & 0.4696 &0.6039 &0.6649& 0.7039 &0.7284\\
     Even & 0.4696& 0.5767 &0.6626& 0.7117& 0.7368 \\
     Odd & 0.4696 &0.6085 &0.6766 &0.7014 &0.7191\\

    \bottomrule
 \end{tabular}%}
\end{table*}

\begin{table*}
  \centering
  \caption{Standard deviation in IoU performance over 10 random view initializations for all policies on the ABC test set in the 3D supervision setting.}
  \label{table:policy_recon_IoU_STD}
  \small

\begin{tabular}[H]{cccccc}
    \toprule

    \multicolumn{1}{c}{} & \multicolumn{5}{c}{\textbf{Number of Images}}  \\
    
     \textbf{Model Setting} & 1 & 2 & 3 & 4 & 5 \\
    \cmidrule[1pt](lr){1-1}\cmidrule[1pt](lr){2-6}
   
     Candidate Policy &0.0813 & \textbf{0.0459}& \textbf{0.0328}& \textbf{0.0256}& \textbf{0.0220}  \\
     Grad - 2 Steps & 0.0813& 0.0519& 0.0398& 0.0318& 0.0267 \\
     Grad - 5 Steps & 0.0813& 0.0495& 0.0381& 0.0309& 0.0273\\
     Grad - 10 Steps & 0.0813& 0.0507& 0.0387& 0.0318& 0.0283\\
     Random & 0.0813& 0.0540& 0.0412& 0.0322& 0.0268\\
     Even & 0.0813& 0.0628& 0.0427& 0.0307& 0.0248\\
     Odd & 0.0813& 0.0530& 0.0383& 0.0336& 0.0296\\

    \bottomrule
 \end{tabular}%}
\end{table*}

\begin{figure}[H]
    \centering
    \includegraphics[width=.8\linewidth]{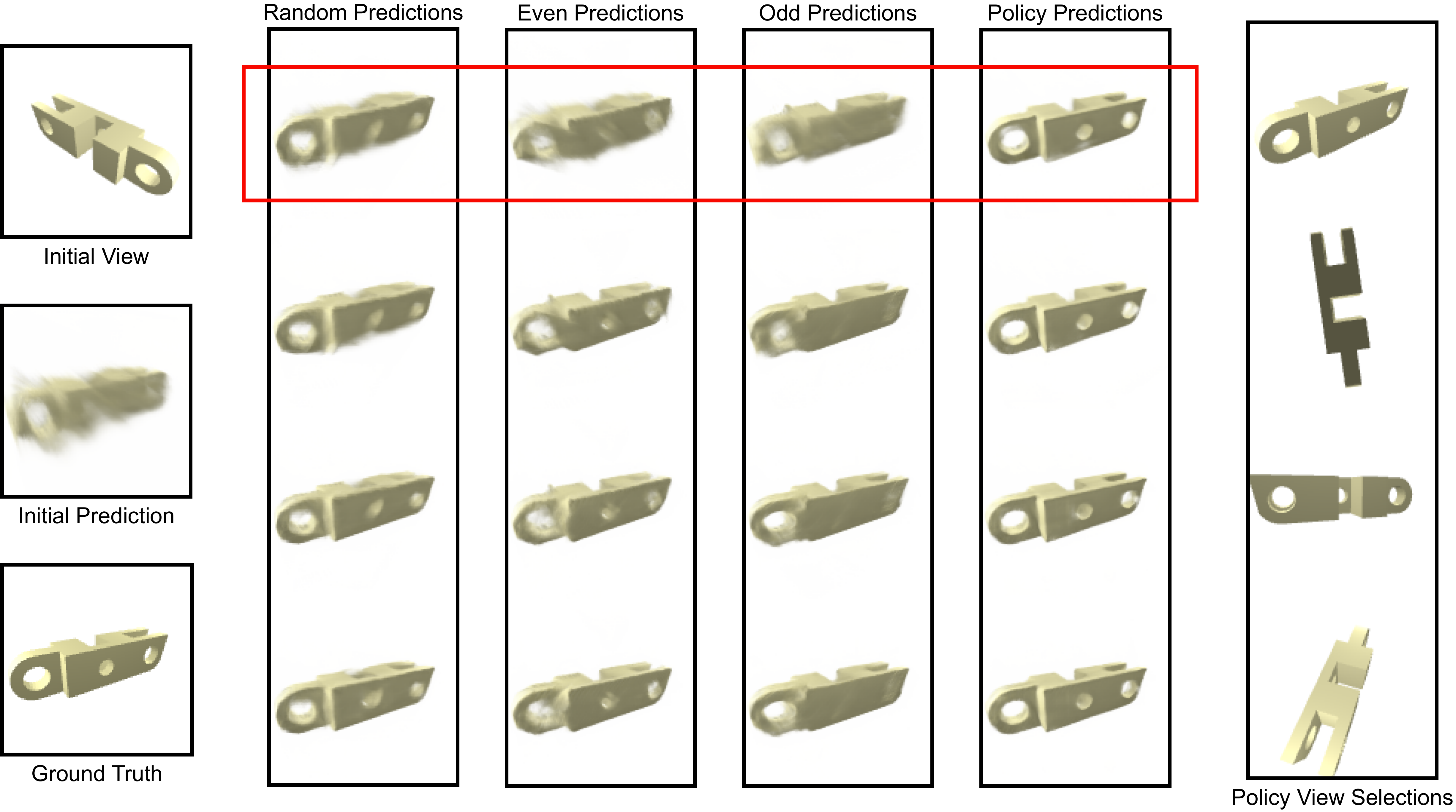}
    \caption{Shape reconstruction comparison over selection policies in the 2D supervision setting, with the \textit{candidate} policy view selections highlighted in final column. Results on ABC. }
    \label{fig:2D_comparision_1}
\end{figure}

\begin{figure}[H]
    \centering
    \includegraphics[width=.8\linewidth]{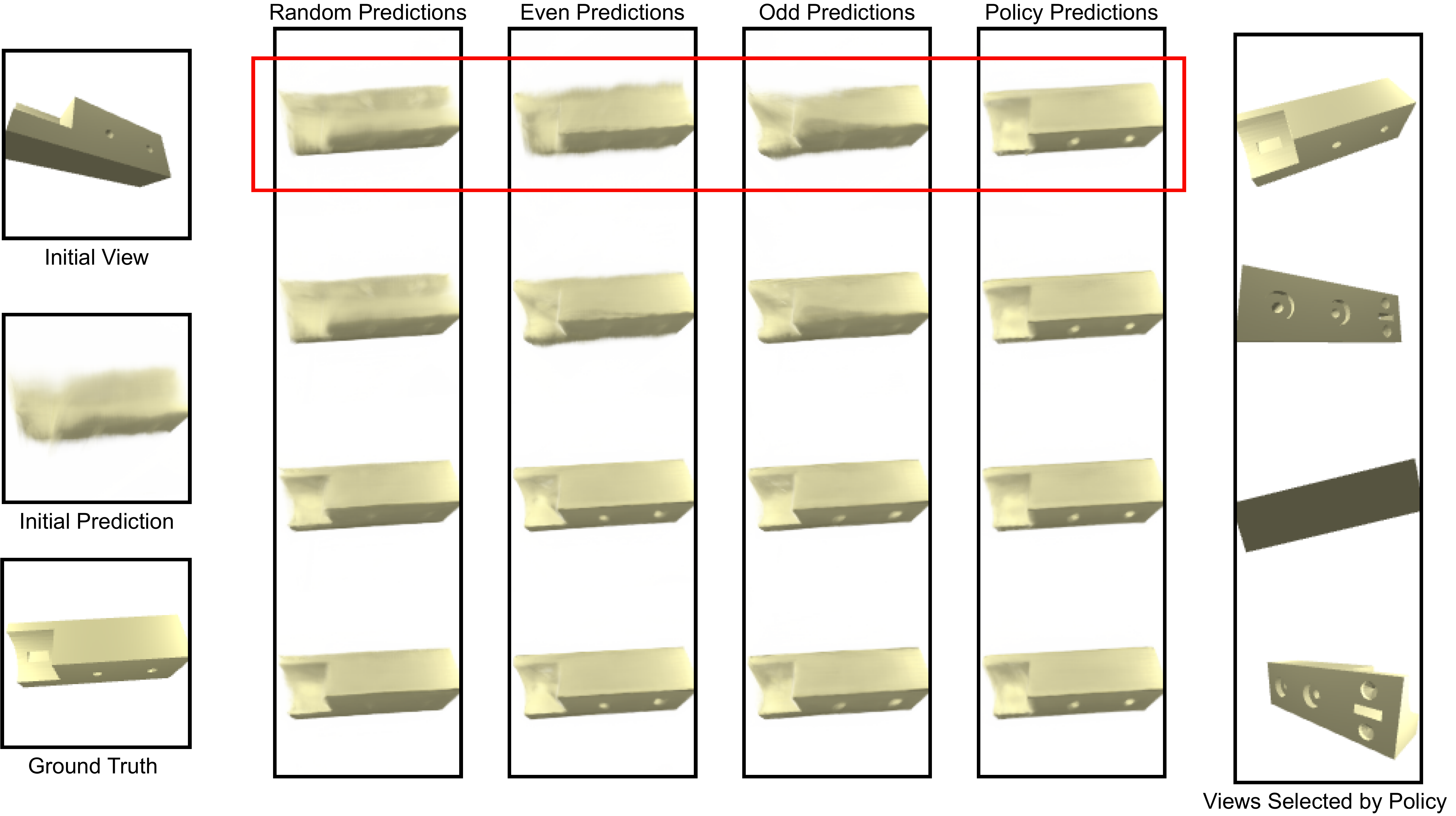}
    \caption{Shape reconstruction comparison over selection policies in the 2D supervision setting, with the \textit{candidate} policy view selections highlighted in final column. Results on ABC. }
    \label{fig:2D_comparision_2}
\end{figure}

\begin{figure}[H]
    \centering
    \includegraphics[width=.8\linewidth]{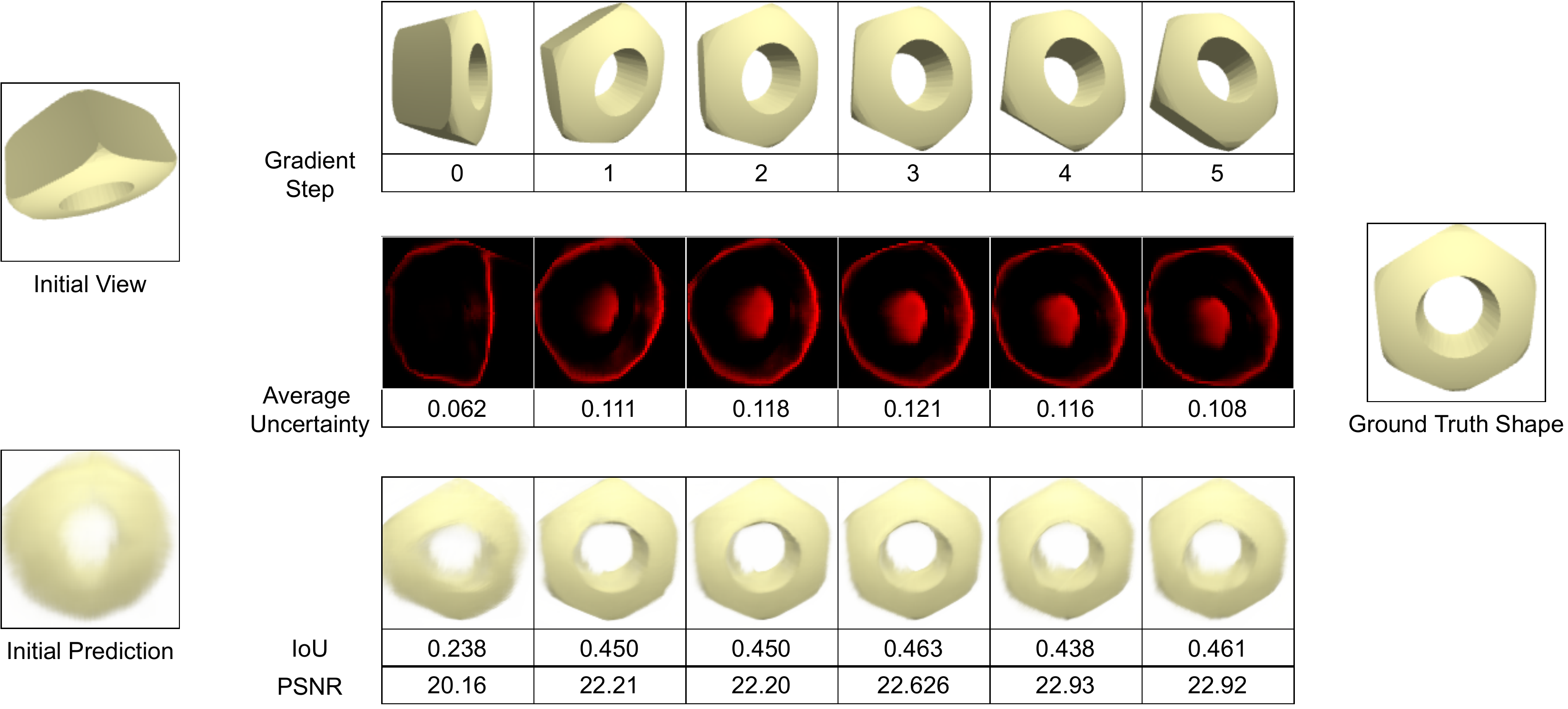}
    \caption{The change in perspective observed in the \textit{gradient} policy gradient updates in the 2D supervision setting, where for each step the current uncertainty, shape prediction, and resulting IoU and PSNR are highlighted. Results on ABC.}
    \label{fig:2D_grad_2}
\end{figure}

We highlight the performance of all policies over IoU and PSNR in the 2D supervision setting on the ABC dataset in Table~\ref{table:policy_recon_IoU_2D} and Table~\ref{table:policy_recon_PSNR_2D}. These evaluations are with 10 random view initializations and over the 500 test set objects. We demonstrate in Table~\ref{table:policy_recon_IoU_worst_2D} and Table~\ref{table:policy_recon_PSNR_worst_2D} the worst case performance of all policies over the 10 random initializations, and show that our \textit{candidate} policy continues to outperform all others. We demonstrate in Table~\ref{table:policy_recon_IoU_STD_2D} and Table~\ref{table:policy_recon_IoU_PSNR_std_2D} standard deviation in IoU of all policies over the 10 random initializations, and show that our \textit{candidate} policy also results in lower variance of performance. In Figures~\ref{fig:2D_comparision_1} and~\ref{fig:2D_comparision_2}, we provide reconstruction results for different policy selection strategies, and in Figure~\ref{fig:2D_grad_2}, we provide additional results for the \textit{gradient} policy's updates over the perspective search space.\looseness-1

\begin{table*}
  \centering
  \caption{IoU performance of all policies on the ABC test set in the 2D supervision setting.}
  \label{table:policy_recon_IoU_2D}
  \small
\begin{tabular}{cccccc}
    \toprule
    \multicolumn{1}{c}{} & \multicolumn{5}{c}{\textbf{Number of Images}}  \\
    \textbf{Model Setting} & 1 & 2 & 3 & 4 & 5 \\
    \cmidrule[1pt](lr){1-1}\cmidrule[1pt](lr){2-6}
   
     Candidate Policy & 0.2996 &	\textbf{0.4436} &	\textbf{0.5062} &	\textbf{0.5419}	& \textbf{0.5644} \\
     Grad - 2 Steps & 0.2996	&0.4228	&0.4846	&0.521	&0.5453 \\
     Grad - 5 Steps & 0.2996	&0.4287	&0.491	&0.527	&0.5496\\
     Grad - 10 & 0.2996	&0.4311	&0.4937&	0.5296&	0.5528\\
     Random & 0.2996&	0.4099&	0.4676&	0.5043	&0.5293\\
     Even & 0.2996&	0.3951&	0.4777&	0.5248&	0.5519\\
     Odd & 0.2996	&0.4218	&0.4887	&0.5202	&0.5438\\
    \bottomrule
 \end{tabular}%}
\end{table*}

\begin{table*}
  \centering
  \caption{Worst case IoU performance of all policies on the ABC test set in the 2D supervision setting.\looseness-1}
  \label{table:policy_recon_IoU_worst_2D}
  \small
\begin{tabular}{cccccc}
    \toprule
    \multicolumn{1}{c}{} & \multicolumn{5}{c}{\textbf{Number of Images}}  \\
    \textbf{Model Setting} & 1 & 2 & 3 & 4 & 5 \\
    \cmidrule[1pt](lr){1-1}\cmidrule[1pt](lr){2-6}
   
     Candidate Policy & 0.1654	& \textbf{0.3505}	&\textbf{0.4305}	&\textbf{0.4784}	&\textbf{0.5088} \\
     Grad - 2 Steps  & 0.1654	&0.3146	&0.393	&0.441	&0.4722 \\
     Grad - 5 Steps & 0.1654	&0.3264	&0.4058	&0.4515	&0.4818\\
     Grad - 10 Steps & 0.1654	&0.3309	&0.4085	&0.4545	&0.4866\\
     Random &0.1654	&0.2952&	0.3677&	0.4165&	0.4515\\
     Even & 0.1654&	0.2407&	0.3745&	0.4537&	0.498 \\
     Odd & 0.1654&	0.3147&	0.4073&	0.4473&	0.4758\\
    \bottomrule
 \end{tabular}%}
\end{table*}

\begin{table*}
  \centering
  \caption{Standard deviation in IoU performance over 10 random view initializations for all policies on the ABC test set in the 2D supervision setting.}
  \label{table:policy_recon_IoU_STD_2D}
  \small
\begin{tabular}{cccccc}
    \toprule
    \multicolumn{1}{c}{} & \multicolumn{5}{c}{\textbf{Number of Images}}  \\
    \textbf{Model Setting} & 1 & 2 & 3 & 4 & 5 \\
    \cmidrule[1pt](lr){1-1}\cmidrule[1pt](lr){2-6}
   
     Candidate Policy  & 0.0749	 & \textbf{0.0531}	 & \textbf{0.0432}	 & \textbf{0.0364}	 & \textbf{0.0322} \\
     Grad - 2 Steps  & 0.0749	 & 0.0612	 & 0.0516	 & 0.0446	 & 0.0458 \\
     Grad - 5 Steps  & 0.0749	 & 0.058	 & 0.0483	 & 0.0422	 & 0.0381\\
     Grad - 10 Steps  & 0.0749	 & 0.057	 & 0.0485	 & 0.0421	 & 0.0373\\
     Random  & 0.0749	 & 0.0642	 & 0.0557	 & 0.0484	 & 0.0435\\
     Even  & 0.0749	 & 0.0795	 & 0.0558	 & 0.0408	 & 0.0326\\
     Odd & 0.0749	 & 0.0599	 & 0.0465	 & 0.0418	 & 0.0893\\
    \bottomrule
 \end{tabular}%}
\end{table*}

\begin{table*}
  \centering
  \caption{PSNR performance of all policies on the ABC test set in the 2D supervision setting.}
  \label{table:policy_recon_PSNR_2D}
  \small
\begin{tabular}{cccccc}
    \toprule
    \multicolumn{1}{c}{} & \multicolumn{5}{c}{\textbf{Number of Images}}  \\
    \textbf{Model Setting} & 1 & 2 & 3 & 4 & 5 \\
    \cmidrule[1pt](lr){1-1}\cmidrule[1pt](lr){2-6}
   
      Candidate Policy & 23.25 	& \textbf{26.23 }	& \textbf{28.04 }	& \textbf{29.26} 	& \textbf{30.04}  \\
     Grad - 2 Steps & 23.25 	& 25.95 	& 27.57 	& 28.65 	& 29.41 \\
     Grad - 5 Steps & 23.25 	& 26.08 	& 27.72 & 28.82 	& 29.55 \\
     Grad - 10 & 23.25 	& 26.12 	& 27.77 &  28.85 	& 29.57 \\
     Random & 23.25 	& 25.74 	& 27.29 	& 28.34 	& 29.12 \\
     Even & 23.25 	& 25.53 	& 27.70 	& 29.13 	& 30.01 \\
     Odd & 23.25 	& 26.02 & 27.85 	& 28.81 	& 29.52 \\
    \bottomrule
 \end{tabular}%}
\end{table*}

\begin{table*}
  \centering
  \caption{Worst case PSNR performance of all policies on the ABC test set in the 2D supervision setting.\looseness-1}
  \label{table:policy_recon_PSNR_worst_2D}
  \small
\begin{tabular}{cccccc}
    \toprule
    \multicolumn{1}{c}{} & \multicolumn{5}{c}{\textbf{Number of Images}}  \\
    \textbf{Model Setting} & 1 & 2 & 3 & 4 & 5 \\
    \cmidrule[1pt](lr){1-1}\cmidrule[1pt](lr){2-6}
   
     Candidate Policy  & 20.38 & \textbf{24.28}	& \textbf{26.32} & \textbf{27.70}	& \textbf{28.70} \\
     Grad - 2 Steps  & 20.38	& 23.72	& 25.58	& 26.75	& 27.66 \\
     Grad - 5 Steps  & 20.38& 23.98	& 25.75	& 27.00	& 27.85\\
     Grad - 10 Steps & 20.38	& 24.05	& 25.86	& 27.09	& 27.88\\
     Random & 20.38	& 23.30	& 25.03	& 26.19	& 27.12\\
     Even & 20.38	& 22.67	& 25.46	& 27.49	& \textbf{28.70}\\
     Odd & 20.38	& 23.75	& 26.04	& 27.08	& 27.91\\
    \bottomrule
 \end{tabular}%}
\end{table*}

\begin{table*}
  \centering
  \caption{Standard deviation in PSNR performance over 10 random view initializations for all policies on the ABC test set in the 2D supervision setting.}
  \label{table:policy_recon_IoU_PSNR_std_2D}
  \small
\begin{tabular}{cccccc}
    \toprule
    \multicolumn{1}{c}{} & \multicolumn{5}{c}{\textbf{Number of Images}}  \\
    \textbf{Model Setting} & 1 & 2 & 3 & 4 & 5 \\
    \cmidrule[1pt](lr){1-1}\cmidrule[1pt](lr){2-6}
   
     Candidate Policy  & 1.459& \textbf{1.058}	& \textbf{0.942}& \textbf{0.832}& 0.731 \\
     Grad - 2 Steps  & 1.459	& 1.191	& 1.081	&1.000	& 0.914 \\
     Grad - 5 Steps  & 1.459	& 1.132	& 1.068	& 0.974	& 0.897\\
     Grad - 10 Steps  & 1.459	& 1.127	& 1.039	& 0.956	& 0.888\\
     Random  & 1.459	& 1.288	& 1.199	& 1.134	& 1.041\\
     Even  & 1.459	& 1.434	& 1.156	& 0.865	& \textbf{0.719}\\
     Odd & 1.459	& 1.197	& 0.991	& 0.934	& 0.856\\
    \bottomrule
 \end{tabular}%}
\end{table*}

\subsection{Occupancy Calibration}
\begin{figure}
     \centering
     \begin{subfigure}[b]{0.32\textwidth}
         \centering
         \includegraphics[width=\textwidth]{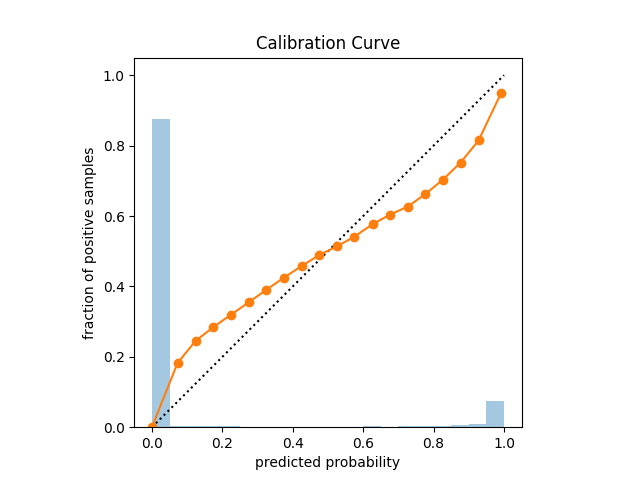}
         \caption{}
         \label{fig:CalCur1}
     \end{subfigure}
     \hfill
     \begin{subfigure}[b]{0.32\textwidth}
         \centering
         \includegraphics[width=\textwidth]{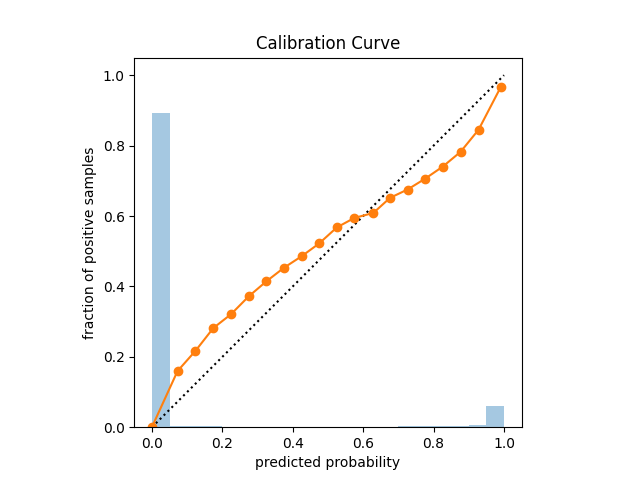}
         \caption{}
         \label{fig:CalCur2}
     \end{subfigure}
     \hfill
     \begin{subfigure}[b]{0.32\textwidth}
         \centering
         \includegraphics[width=\textwidth]{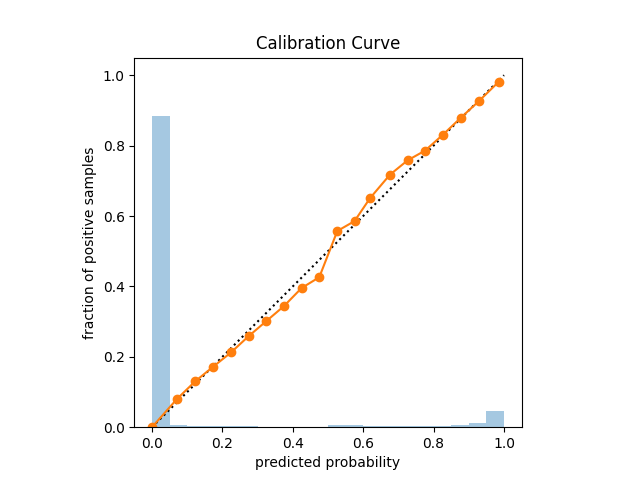}
         \caption{}
         \label{fig:CalCur3}
     \end{subfigure}
        \caption{Calibration curves on a) the validation set, b) the test set, and c) the test set after exponential smoothing of predicted occupancy probabilities with $\beta$ = 0.7. This is performed with 20 bins and the blue columns for each bin indicates there relative. Results on ABC.}
        \label{fig:three graphs}
\end{figure}

We define the uncertainty of occupancy predictions using the direct predicted probabilities in Equation ~\ref{eq:occ_uncert}, however it is well established that classification probabilities can be poorly calibrated with accuracy~\citep{guo2017calibration}. We demonstrate that this is also true for our model, by plotting the validation set calibration curve of our reconstruction model in the 3D supervision setting on the ABC dataset in Figure~\ref{fig:CalCur1}. This posses an average $l_1$ calibration error over 20 bins of 0.073\footnote{We highlight average calibration error instead of expected calibration error due to the dramatically uneven number of samples across bins.}. However, we suggest that the hyper-parameter $\beta$ in this equation can act as a re-calibration tool of the predicted probabilities: 

\[   
\hat{o}(t) = 
     \begin{cases}
       t \leq .5 & \frac{1+\left(2\left|o(t)-.5\right|\right)^{\beta}}{2} \\
       t \geq .5  & \frac{1-\left(2\left|o(t)-.5\right|\right)^{\beta}}{2}
     \end{cases}
\]

To demonstrate this, we compute expected calibration curves over the validation set for different values of $\beta$ in Table~\ref{table:calibration}. From this we see that the calibration error can be drastically reduced by a simple exponential scaling with values 0.7 of the predicted occupancy. To demonstrate that this trend continues to the test set, we plot the calibration curves before and after scaling in Figures~\ref{fig:CalCur2} and~\ref{fig:CalCur3}, where the average calibration error reduces from 0.064 to 0.017.

\begin{table*}
  \centering
  \caption{Average calibration error over 20 bins for different values of $\beta$ for the ABC validation set.}
  \label{table:calibration}
  \small
\begin{tabular}{c ccccc ccccc ccccc ccccc }
    \toprule
      $\beta$  & 0.5 &  0.6 &  0.7 &  0.8 &  0.9 &  1.0 &  1.1 &  1.2 &  1.3  &  1.4 \\
    \cmidrule[1pt](lr){1-1}\cmidrule[1pt](lr){2-11}
   
      Calibration Error  & 0.056 &  0.031 & 0.023 &  0.035 & 0.052 & 0.073 & 0.094 & 0.119 &  0.136  & 0.157 \\

    \bottomrule
 \end{tabular}%}
\end{table*}

\begin{table*}
  \centering
  \caption{Architecture for CNN in the 3D supervision setting.}
  \vspace{0.2cm}
  \label{table:3D_Network}
  \scalebox{.75}{
  \begin{tabular}{|c|c|c|c|}
  \toprule
    Index & Input & Operation & Output Shape \\

    \midrule
    (1)& Input & Conv (3$\times$ 3) + BN + ReLU  & 3 $\times$ 128 $\times$ 128 \\
    (2)& (1) & Conv (3$\times$ 3) + BN + ReLU  & 32 $\times$ 64 $\times$ 64 \\
    (3)& (2) & Conv (3$\times$ 3) + BN + ReLU  & 32 $\times$ 64 $\times$ 64\\
    (4)& (3) & Conv (3$\times$ 3) + BN + ReLU  & 32 $\times$ 64 $\times$ 64\ \\
    (5)& (4) & Conv (3$\times$ 3) + BN + ReLU  & 32 $\times$ 64 $\times$ 64\ \\
    (6)& (5) & Conv (3$\times$ 3) + BN + ReLU  & 32 $\times$ 64 $\times$ 64\ \\
    (7)& (6) & Conv (3$\times$ 3) + BN + ReLU  & 32 $\times$ 64 $\times$ 64\ \\
    (8)& (7) & Conv (3$\times$ 3) + BN + ReLU  & 64 $\times$ 32 $\times$ 32\ \\
    (9)& (8) & Conv (3$\times$ 3) + BN + ReLU  & 64 $\times$ 32 $\times$ 32 \\
    (10)& (8) & Conv (3$\times$ 3) + BN + ReLU  & 64 $\times$ 32 $\times$ 32 \\
    (11)& (10) & Conv (3$\times$ 3) + BN + ReLU  & 64 $\times$ 32 $\times$ 32 \\
    (12)& (11) & Conv (3$\times$ 3) + BN + ReLU  & 64 $\times$ 32 $\times$ 32 \\
    (13)& (12) & Conv (3$\times$ 3) + BN + ReLU  & 64 $\times$ 32 $\times$ 32 \\
    (14)& (13) & Conv (3$\times$ 3) + BN + ReLU  & 128 $\times$ 16 $\times$ 16 \\
    (15)& (14) & Conv (3$\times$ 3) + BN + ReLU  & 128 $\times$ 16 $\times$ 16 \\
    (16) &(15) & Conv (3$\times$ 3) + BN + ReLU  &128 $\times$ 16 $\times$ 16 \\
    (17) & (16) & Conv (3$\times$ 3) + BN + ReLU  & 128 $\times$ 16 $\times$ 16\\
    (18) & (17) & Conv (3$\times$ 3) + BN + ReLU  & 128 $\times$ 16 $\times$ 16\\
    (19) & (18) & Conv (3$\times$ 3) + BN + ReLU  & 128 $\times$ 16 $\times$ 16 \\
    (20) & (19) & Conv (3$\times$ 3) + BN + ReLU  & 256 $\times$ 8 $\times$ 8\\
    (21) & (20) & Conv (3$\times$ 3) + BN + ReLU  &  256 $\times$ 8 $\times$ 8\\
    (22) & (21) & Conv (3$\times$ 3) + BN + ReLU  & 256 $\times$ 8 $\times$ 8\\
    (23) & (22) & Conv (3$\times$ 3) + BN + ReLU  &  256 $\times$ 8 $\times$ 8\\
    (24) & (23) & Conv (3$\times$ 3) + BN + ReLU  &  256 $\times$ 8 $\times$ 8\\
    (25) & (24) & Conv (3$\times$ 3) + BN + ReLU  &  256 $\times$ 8 $\times$ 8\\
    
  \bottomrule
 
  \end{tabular}}
\end{table*}
\begin{table*}
  \centering
  \caption{Architecture for CNN in the 2D supervision setting.}
  \vspace{0.2cm}
  \label{table:2D_Network}
  \scalebox{.75}{
  \begin{tabular}{|c|c|c|c|}
  \toprule
    Index & Input & Operation & Output Shape \\

    \midrule
    (1)& Input & Conv (3$\times$ 3) + BN + ReLU  & 3 $\times$ 128 $\times$ 128 \\
    (2)& (1) & Conv (3$\times$ 3) + BN + ReLU  & 40 $\times$ 64 $\times$ 64 \\
    (3)& (2) & Conv (3$\times$ 3) + BN + ReLU  & 40 $\times$ 64 $\times$ 64\\
    (4)& (3) & Conv (3$\times$ 3) + BN + ReLU  & 40 $\times$ 64 $\times$ 64\ \\
    (5)& (4) & Conv (3$\times$ 3) + BN + ReLU  & 40 $\times$ 64 $\times$ 64\ \\
    (6)& (5) & Conv (3$\times$ 3) + BN + ReLU  & 40 $\times$ 64 $\times$ 64\ \\
    (7)& (6) & Conv (3$\times$ 3) + BN + ReLU  & 40 $\times$ 64 $\times$ 64\ \\
    (8)& (7) & Conv (3$\times$ 3) + BN + ReLU  & 80 $\times$ 32 $\times$ 32\ \\
    (9)& (8) & Conv (3$\times$ 3) + BN + ReLU  & 80 $\times$ 32 $\times$ 32 \\
    (10)& (8) & Conv (3$\times$ 3) + BN + ReLU  & 80 $\times$ 32 $\times$ 32 \\
    (11)& (10) & Conv (3$\times$ 3) + BN + ReLU  & 80 $\times$ 32 $\times$ 32 \\
    (12)& (11) & Conv (3$\times$ 3) + BN + ReLU  & 80 $\times$ 32 $\times$ 32 \\
    (13)& (12) & Conv (3$\times$ 3) + BN + ReLU  & 80 $\times$ 32 $\times$ 32 \\
    (14)& (13) & Conv (3$\times$ 3) + BN + ReLU  & 160 $\times$ 16 $\times$ 16 \\
    (15)& (14) & Conv (3$\times$ 3) + BN + ReLU  & 160 $\times$ 16 $\times$ 16 \\
    (16) &(15) & Conv (3$\times$ 3) + BN + ReLU  & 160 $\times$ 16 $\times$ 16 \\
    (17) & (16) & Conv (3$\times$ 3) + BN + ReLU  & 160 $\times$ 16 $\times$ 16\\
    (18) & (17) & Conv (3$\times$ 3) + BN + ReLU  & 160 $\times$ 16 $\times$ 16\\
    (19) & (18) & Conv (3$\times$ 3) + BN + ReLU  & 160 $\times$ 16 $\times$ 16 \\
    (20) & (19) & Conv (3$\times$ 3) + BN + ReLU  & 320 $\times$ 8 $\times$ 8\\
    (21) & (20) & Conv (3$\times$ 3) + BN + ReLU  &  320 $\times$ 8 $\times$ 8\\
    (22) & (21) & Conv (3$\times$ 3) + BN + ReLU  & 320 $\times$ 8 $\times$ 8\\
    (23) & (22) & Conv (3$\times$ 3) + BN + ReLU  &  320 $\times$ 8 $\times$ 8\\
    (24) & (23) & Conv (3$\times$ 3) + BN + ReLU  &  320 $\times$ 8 $\times$ 8\\
    (25) & (24) & Conv (3$\times$ 3) + BN + ReLU  &  320 $\times$ 8 $\times$ 8\\
    
  \bottomrule
 
  \end{tabular}}
\end{table*}

\end{document}